\documentclass[sigconf]{aamas}  
\usepackage{balance}  

\usepackage{booktabs}

\setcopyright{ifaamas}  
\acmDOI{}  
\acmISBN{}  
\acmConference[AAMAS'19]{Proc.\@ of the 18th International Conference on Autonomous Agents and Multiagent Systems (AAMAS 2019)}{May 13--17, 2019}{Montreal, Canada}{N.~Agmon, M.~E.~Taylor, E.~Elkind, M.~Veloso (eds.)}  
\acmYear{2019}  
\copyrightyear{2019}  
\acmPrice{}  

\usepackage[utf8]{inputenc} 
\usepackage[T1]{fontenc}    
\usepackage{hyperref}       
\usepackage{url}            
\usepackage{booktabs}       
\usepackage{amsfonts}       
\usepackage{nicefrac}       
\usepackage{microtype}      

\usepackage{latexsym}
\usepackage{amsmath}
\usepackage{amsthm}
\usepackage{amsfonts}
\usepackage{graphicx}
\usepackage{epstopdf}
\usepackage{multirow}
\usepackage{csquotes}
\usepackage[]{algorithm2e}
\usepackage{longtable}
\usepackage{enumitem}
\usepackage{rotating}
\usepackage{bbm}
\usepackage{lscape}
\usepackage{wrapfig}
\usepackage{subcaption}
\usepackage{nicefrac}       
\usepackage{microtype}      
\usepackage{framed}
\usepackage{tabularx}
\usepackage{booktabs}
\usepackage{tikz}
\usepackage{pifont}
\usepackage{centernot}

\newcommand{\cmark}{\ding{51}}%
\newcommand{\xmark}{\ding{55}}%

\newcommand{\pol}[0]{\pmb{\pi}}
\newcommand\independent{\protect\mathpalette{\protect\independenT}{\perp}}
\def\independenT#1#2{\mathrel{\rlap{$#1#2$}\mkern2mu{#1#2}}}
\usepackage[makeroom]{cancel}

\begin{document}

\title{On the Pitfalls of Measuring Emergent Communication}




\author{Ryan Lowe}  
\affiliation{
\institution{Mila \& Facebook AI Research}
\city{Montreal}
\state{QC}
}
\email{ryan.lowe@cs.mcgill.ca}

\author{Jakob Foerster}
\affiliation{
\institution{Facebook AI Research}
\city{Menlo Park}
\state{CA}
}

\author{Y-Lan Boureau}  
\affiliation{
\institution{Facebook AI Research}
\city{New York}
\state{NY}
}

\author{Joelle Pineau}  
\affiliation{
\institution{Mila \& Facebook AI Research}
\city{Montreal}
\state{QC}
}

\author{Yann Dauphin}  
\affiliation{
\institution{Google AI}
\city{Accra}
\country{Ghana}
}


\begin{abstract}  
How do we know if communication is emerging in a multi-agent system? 
The vast majority of recent papers on emergent communication show that adding a communication channel leads to an increase in reward or task success.
This is a useful indicator, but provides only a coarse measure of the agent's learned communication abilities.
As we move towards more complex environments, it becomes imperative to have a set of finer tools that allow qualitative and quantitative insights into the emergence of communication. This may be especially useful to allow humans to monitor agents' behaviour, whether for fault detection, assessing performance, or even building trust. 
In this paper, we examine a few intuitive existing metrics for measuring communication, and show that they can be misleading. 
Specifically, by training deep reinforcement learning agents to play simple matrix games augmented with a communication channel, we find a scenario where agents appear to communicate (their messages provide information about their subsequent action), and yet the messages do not impact the environment or other agent in any way. We explain this phenomenon using ablation studies and by visualizing the representations of the learned policies.  We also survey some commonly used metrics for measuring emergent communication, and provide recommendations as to when these metrics should be used.
\end{abstract}

\keywords{Deep learning; Learning agent capabilities; Multi-agent learning}  

\maketitle

\section{Introduction}


Communication through language is one of the hallmarks of human intelligence; it allows us to share information efficiently between humans and coordinate on shared tasks. This is a strong motivation to develop machines that can communicate, so that they can share their knowledge with us and help us coordinate on tasks where humans alone perform poorly. 
One of the approaches for developing agents capable of language is through \textit{emergent communication}. The idea is to use an environment consisting of randomly initialized agents and a dedicated communication channel, which the agents can learn to use to accomplish tasks that require coordination. This approach is motivated by a utilitarian view of communication, where language is viewed as one of many tools an agent can use towards achieving its goals, and an agent `understands' language once it can use it to accomplish these goals \citep{wittgenstein1954,gauthier2016paradigm}.  The question of emergent communication is also of scientific interest, as it could provide insights into the origins of human language \cite{nowak1999evolution}. 

Recently, there has been a significant revival of emergent communication research using methods from \textit{deep reinforcement learning} (deep RL). In this line of work, researchers define a multi-agent environment where agents possess a (usually discrete and costless) communication channel, and train the agents' policies (represented by deep neural networks) to maximize the cumulative reward. 
In most cases, the multi-agent environment is a cooperative \textit{referential game}, where one agent (the `speaker') has some private information about a public observation, and must communicate it accurately to the other agent (the `listener') \cite{das2017learning,lazaridou2018emergence,kottur2017natural,choi2018compositional,lazaridou2016multi,havrylov2017emergence,evtimova2017emergent}. In other works, agents navigate a simulated world with a limited field of view, and communication can help the agents coordinate to achieve a common goal \cite{sukhbaatar2016learning,jaques2018intrinsic,foerster2018bayesian,bogin2018emergence}. Algorithmic development has led to emergent communication in games with pixel-based inputs \cite{lazaridou2018emergence,choi2018compositional,bogin2018emergence}, in settings incorporating human language \cite{lazaridou2016multi,lee2017emergent}, and even in more complex games that require inferring the belief state of the opponent, like Hanabi \cite{foerster2018bayesian}.

Despite these advances, there has been no formal study evaluating \textit{how} we measure emergent communication. 
Most papers in this area show that adding a communication channel helps the agents achieve a higher reward, and attempt to understand the communication protocols qualitatively by, for example, plotting the distribution over messages in various states of the environment. This is a reasonable approach: if adding a communication channel increases the reward in an environment, then the agents are making some use of that channel. 
But it is useful to quantify more finely the degree of communication as it can provide more insights into agent behaviour. This may be essential for human monitoring; we want to understand why agents are making certain decisions, and understanding their language would make this significantly easier. 
Our aim is to develop a set of evaluation tools for detecting and measuring emergent communication that increases the robustness of emergent communication research on the way to developing agents that communicate with humans.


In this paper, we take a step in this direction by analyzing some intuitive metrics and tests that have been used in the recent literature for measuring emergent communication. In particular, we show that some of these metrics, including speaker consistency \cite{jaques2018intrinsic} and simply examining the agents' policies qualitatively, can be misleading if not properly understood.
We use a simple class of matrix games augmented with a bidirectional communication channel, which we call Matrix Communication Games (MCGs), to study the emergence of communication among deep RL agents. MCGs 
are simple and efficient for learning, and have the appealing property that the resulting communication policies are interpretable relative to more complex gridworld environments. 

To aid in our analysis, 
we categorize emergent communication metrics into two broad classes: 
those that measure \textit{positive signaling}, which indicates that an agent is sending messages that are related in some way with its observation or action; and those that measure \textit{positive listening}, indicating that the messages are influencing the agents' behaviour in some way. Both positive signaling and positive listening are desirable; we want to develop communicative agents that can both speak about their world and intentions, and can interpret the messages of others and respond accordingly. While intuitively it might seem that positive listening is a prerequisite for positive signaling (otherwise, why would the agent signal in the first place?), we show that, surprisingly, this is not always true for deep RL agents; it is possible for positive signaling to emerge as a byproduct of the task and policy architecture. 



Our main contributions are as follows. 
\begin{itemize}
\item We briefly survey the existing metrics for detecting emergent communication in the deep RL community, and find that the majority measure positive signaling rather than positive listening. 
\item We run experiments on MCGs and show that agents can exhibit positive signaling without positive listening. This lack of positive listening falsifies the hypothesis that the agents are communicating to share information to increase the reward of both agents. Thus, positive listening is an important tool for understanding the purpose of emergent communication.   
\item We propose a metric, the causal influence of communication (CIC), designed to directly measure positive listening. Note that a variant of CIC was developed concurrently by \cite{jaques2018intrinsic}, where they showed that agents trained with causal influence as an additional reward performed better in several multi-agent settings. We adopt their terminology in this paper. 
\item We discuss other metrics and tests for measuring emergent communication, and make recommendations for future work in this area.\footnote{Code available at \texttt{github.com/facebookresearch/measuring-emergent-comm}.} 
\end{itemize}

\section{Background}

\subsection{Markov games}

The MCG environment in this paper can be considered a multi-agent extension of partially observable Markov decision processes (POMDPs) called partially observable Markov games \cite{littman1994markov}. 
A Markov game for $N$ agents is defined by a set of states $\mathcal{S}$, $N$ sets of actions $\mathcal{A}_1,...,\mathcal{A}_N$ and $N$ sets of observations $\mathcal{O}_1,...,\mathcal{O}_N$, one each per agent. To choose actions, each agent $i$ uses a stochastic policy $\pol_{\theta_i} : \mathcal{O}_i \times \mathcal{A}_i \mapsto [0,1]$, which produces the next state according to the state transition function $\mathcal{T} : \mathcal{S} \times \mathcal{A}_1 \times \cdots \times \mathcal{A}_N \mapsto \mathcal{S}$.
Each agent $i$ obtains rewards as a function of the state and agent's action $r_i : \mathcal{S} \times \mathcal{A}_1 \times \cdots \times \mathcal{A}_N \mapsto \mathbb{R}$, and receives a private observation correlated with the state $o_i : \mathcal{S} \mapsto \mathcal{O}_i$. The initial states are determined by a distribution $\rho : \mathcal{S} \mapsto [0,1]$. Each agent $i$ aims to maximize its own total expected return $R_i = \sum_{t=0}^T \gamma^t r^t_i$ where $\gamma$ is a discount factor and $T$ is the time horizon.

More specifically, in the games we consider the action space $\mathcal{A}_i$ for each agent $i$ can be subdivided into disjoint \textit{environment actions} 
$\mathcal{A}^e_i$, and \textit{communication actions} $\mathcal{A}^m_i$, such that $\mathcal{A}^e_i \cup \mathcal{A}^m_i = \mathcal{A}_i$ and $\mathcal{A}^e_i \cap \mathcal{A}^m_i = \emptyset$. Environment actions are those that have a direct effect on the environment dynamics and the rewards obtained by the agent. 
We model each agent's communication action as a sequence of discrete symbols sent over a dedicated communication channel, which are observed by the other agents at the next time step.
Communication actions do not affect the environment dynamics (other than being observed by the other agent), and incur a fixed cost $r_c \in \mathbb{R}_{<0}$.
In our paper, we consider the \textit{cheap talk} setting \citep{farrell1996cheap} where $r_c = 0$, and leave an examination of costly signaling (e.g. \citep{zahavi1975mate, gintis2001costly}) to future work.

\subsection{Policy gradient methods}

Policy gradient methods are a popular choice for a variety of RL tasks.
The main idea is to directly adjust the parameters $\theta$ of the policy in order to maximize the objective $J(\theta) = \mathbb{E}_{s \sim p^{\pol}, a \sim {\pol}_\theta}[R]$ by taking steps in the direction of $\nabla_\theta J(\theta)$. The gradient of the policy can be written as \cite{sutton2000policy}:
\begin{equation}
\label{eq:reinforce}
\nabla_\theta J(\theta) = \mathbb{E}_{s \sim p^{\pol}, a \sim {\pol}_\theta} [\nabla_\theta \log \pol_\theta(a|s) Q^{\pol} (s,a)],
\end{equation}
where $p^{\pol}$ is the state distribution, and $Q^{\pol} (s,a)$ is an estimate of the expected value of taking action $a$ in state $s$. 

The policy gradient theorem has given rise to several practical algorithms, which often differ in how they estimate $Q^{\pol}$. In the simplest case, one can use a sample return $R^t = \sum_{i=t}^T \gamma^{i-t}r_i$, which leads to the REINFORCE algorithm \cite{williams1992simple}. We use the REINFORCE algorithm in our work, as it is simple, and has been by far the most popular approach among emergent communication works using deep RL \cite{sukhbaatar2016learning,das2017learning,lazaridou2016multi,kottur2017natural,cao2018emergent,lazaridou2018emergence,jaques2018intrinsic,havrylov2017emergence}.

\section{Emergent communication metrics}
\label{sec:metrics}

\subsection{A categorization of metrics}


When analyzing metrics to measure a certain quantity, it is important to ask what that quantity actually represents. What does it mean for agents to be communicating with each other in a reinforcement learning setting? We take a pragmatic perspective, and identify two broad prerequisites for communication to occur: (1) one agent needs to produce a signal that is in some way correlated with its observation or intended action, and (2) another agent needs to update its belief or alter its behaviour after observing the signal. 

We define \textit{positive signaling} as behaviour that satisfies criterion (1), and \textit{positive listening} as behaviour that satisfies criterion (2). 
To formalize these intuitions, we provide very broad definitions these terms below. 

\begin{definition}[Positive Signaling] Let $\bar{m} = (m_0, m_1, ..., m_T)$ be the sequence of messages sent by an agent over the course of a trajectory of length $T$, and similarly for $\bar{o}=(o_0, o_1, ..., o_t)$ and $\bar{a}=(a_0, a_1, ..., a_T)$. An RL agent exhibits positive signaling if either $\bar{m} 
\centernot\independent{} \bar{o}$ or $\bar{m} 
\centernot\independent{} \bar{a}$, i.e.\@ if $\bar{m}$ is statistically dependent (indicated by $\centernot\independent{}$) in some way with either $\bar{a}$ or $\bar{o}$. 
\end{definition}

\begin{definition}[Positive Listening] An RL agent exhibits positive listening if there exists a message generated by another agent $m \in \mathcal{A}^m_i$, for some $i \in \{1, ..., N\}$ such that $||\pol(o, \mathbf{0}) - \pol(o, m)||_\tau > 0$, where $\mathbf{0}$ is the 0 vector, and $||\cdot ||_\tau$ is a distance in the space of expected trajectories followed by $\pol$. 
\end{definition}

Evidently, these definitions are very loose, and most agents in an multi-agent environment with the capacity to communicate will satisfy them to some degree. 
However, we can speak of the degree or extent to which an agent exhibits positive signaling or positive listening behaviour, and measure this using metrics or tests. Thus, these terms are useful for categorizing different metrics of emergent communication.

\subsection{What metrics are being used now?}

We now conduct a brief review of the existing metrics being used in papers applying deep reinforcement learning to the problem of emergent communication, and categorize them as being indicative of positive signaling or positive listening. 
We focus on metrics that are the most prevalent in the recent deep RL literature to give an overview of how current research in this area is being conducted. 
From our review, we find that only one metric (instantaneous coordination) is explicitly designed to measure positive listening, and that it has several shortcomings. This motivates our definition of a new metric, the causal influence of communication, in Section \ref{sec:cic}.

\paragraph{Reward and task completion}  
As previously mentioned, in all papers we surveyed, the authors either measure the task completion \% of their proposed algorithm (for referential games \cite{lazaridou2016multi,lazaridou2018emergence,evtimova2017emergent}), or show that adding a communication channel increases the agents' reward (for games where the agents interact with the world \cite{bogin2018emergence,choi2018compositional,sukhbaatar2016learning}). For referential games \cite{das2017learning,lazaridou2018emergence}, this is an adequate measure of communicative performance because these games are \textit{non-situated}. In the terminology of \cite{wagner2003progress}, a non-situated environment (or simulation) is one where agents' actions consist solely of sending and receiving signals. Non-embodied agents in these environments do not have non-communicative actions that affect other objects or each other. Thus, if task success increases in this setting, it is likely because the sender agent has developed a better communication protocol, or the listener agent has become better at understanding it. 

\begin{sloppypar}
In \textit{situated} environments \cite{wagner2003progress}, where agents have non-communicative actions that affect the environment and/or modify their internal state, an increase in reward is still a loose measure of both positive listening and positive signaling. If adding a communication channel leads to an increase in reward, then agents must be using that channel to transmit some kind of information that is affecting their behaviour. However, when comparing multiple algorithms on the same environment, an increase in reward may not necessarily indicate improved communication of one algorithm over another, as this increase could be simply due to the improved action policies. Thus, in this paper we advocate for more fine-grained metrics for measuring the quality of emergent communication protocols. One of the reasons for our adoption of MCGs is that they are in some ways the simplest situated environment (agents have non-communicative interactions that affect each other's reward). 
\end{sloppypar}

\paragraph{Qualitative analysis of messages given states}
An equally common practice is to analyze the communication policies qualitatively, to interpret what the agents are saying. This is done most commonly by visualizing which messages were sent with which inputs or observations \cite{sukhbaatar2016learning,lazaridou2016multi,havrylov2017emergence,jaques2018intrinsic,bogin2018emergence}. For very simple environments, such as the riddle game \cite{foerster16a}, the authors are able to make a full tree describing the agents' policies for all states in the game. However, for other games with larger state spaces, papers simply show some messages and inputs that the authors observed to have high co-occurrence. While this is a useful practice to understand the agents' behaviour, it is usually only testing positive signaling, as it does not analyze how the message changes the other agent's behaviour.

\paragraph{Speaker consistency}
Some papers quantify the degree of alignment between an agent's messages and its actions \cite{bogin2018emergence,jaques2018intrinsic}. One such example is the \textit{speaker consistency} (SC), proposed in \cite{jaques2018intrinsic}. The easiest way to understand the speaker consistency is as the mutual information between an agent's message and its future action:
\begin{equation}
\label{eq:sc}
SC = \sum_{a \in \mathcal{A}^e} \sum_{m \in \mathcal{A}^m} p(a, m) \log \frac{p(a, m)}{p(a)p(m)}
\end{equation}
where the probabilities $p(a, m) = \frac{1}{N} \sum_{i =1}^N \mathbbm{1}_{\{\text{act}=a, \text{comm}=m\}}$ are calculated empirically by averaging (message, action) co-occurrences over the $N$ episodes in each epoch.\footnote{The SC metric in \cite{jaques2018intrinsic} is a normalized version of the metric shown here.} We provide an algorithm for calculating SC in the Appendix. 

The speaker consistency is a measure of positive signaling, as it indicates a statistical relationship between the messages and actions of an agent, and does not tell us anything about how these messages are interpreted. 
On the surface, SC is a useful measure of communication because it tells us how much an agent's message reduces the uncertainty about its subsequent action.
Importantly, when an agent learns a deterministic policy independent of observations, the SC will be 0. Because of these appealing properties, we primarily focus on SC as our metric of comparison in our experiments.

\paragraph{Context independence}
Context independence (CI), introduced by \cite{bogin2018emergence}, is designed to measure the degree of alignment between an agent's messages and task concepts (corresponding to a particular part of the input space). 
It is calculated as follows:
\begin{align*}
\forall c: \hspace{2mm} m^c = \arg \max_m p_{cm}(c|m) \\
CI(p_{cm},p_{mc}) = \frac{1}{|\mathcal{C}|}\sum_c p_{mc} (m^c|c) \cdot p_{cm}(c|m^c)
\end{align*}
where $\mathcal{C}$ is the space of all concepts, $p_{cm}(c|m)$ is the conditional probability of a concept given a message, and similarly for $p_{mc}(m|c)$. This quantity relies on having a well-defined notion of `concept' (in \cite{bogin2018emergence}, this corresponds e.g.\@ to the number and colour of objects), and a way of estimating $p_{cm}$ and $p_{mc}$ (\cite{bogin2018emergence} use an IBM model 1 \cite{brown1993mathematics}).

Context independence captures the same intuition as speaker consistency: if a speaker is consistently using a specific word to refer to a specific concept, then communication has most likely emerged. Thus, it is also a measure of positive signaling. The difference is that CI emphasizes that a single symbol should represent a certain concept or input, whereas a high speaker consistency can be obtained using a set of symbols for each input, so long as this set of symbols is (roughly) disjoint for different inputs. 

\begin{figure*}
    \centering
    \includegraphics[width=0.9\linewidth]{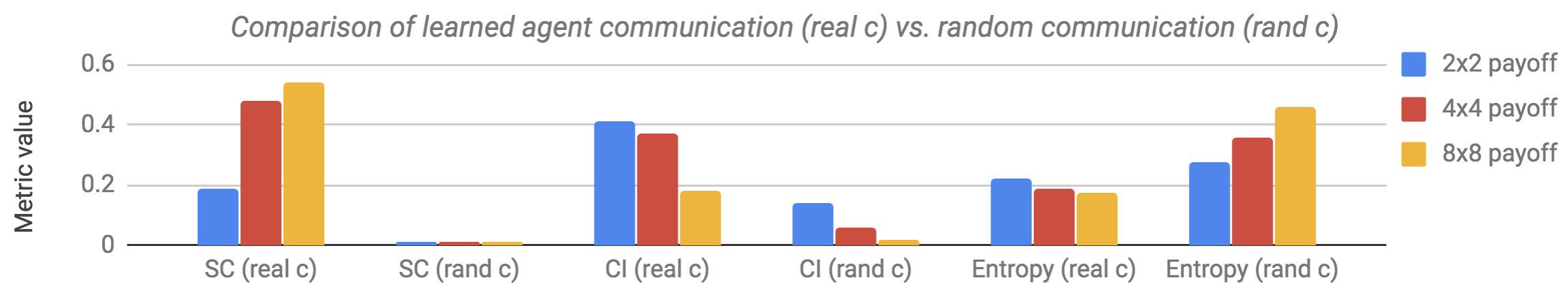}
    \vspace{-3mm}
    \caption{When comparing the speaker consistency (SC), context independence (CI), and entropy of the message policy (divided by 5) between agents who learned to signal (real $c$) and agents who message randomly (rand $c$), it appears as though the agents are communicating (note we expect a lower entropy for a communicating policy). }
    \label{fig:bar_chart}
\end{figure*}

\paragraph{Entropy of message distribution}
Another common practice is to measure the perplexity or entropy ($H(\pol^c)$) of the distribution over messages for different inputs \cite{havrylov2017emergence,choi2018compositional,evtimova2017emergent}. Different papers interpret the meaning of this quantity differently, but generally if the message distribution has low entropy for a given input, then the speaker is consistently using the same message to describe that input. This differs from the speaker consistency metric, as it does not measure whether or not the speaker is using a \textit{different} message for each input. The entropy actually measures neither positive signaling (an agent that always outputs the same message, independent of observation and action, will have low entropy), nor positive listening (it does not take into account the other agent's response to the messages), which gives it questionable utility as a metric.

\paragraph{Instantaneous coordination}
Another metric considered in \cite{jaques2018intrinsic} is instantaneous coordination (IC). 
IC uses the same formula as for SC (Eq. \ref{eq:sc}), except it calculates the mutual information between one agent's message and the \textit{other} agent's action, again by averaging (message, action) co-occurrences over episodes (see Appendix for algorithm). IC is a measure of positive listening; however, because of the way $p(a, m)$ is calculated, it only detects cases where a message from one agent changes the other agent's action regardless of context. 

To illustrate why this is undesirable, we describe a simple matrix communication game (MCG) example (see Section \ref{sec:mcgs} for description) where IC would fail to detect positive listening. Let us draw the entries of the payoff matrix $R \in \mathbb{R}^{2 \times 2}$ from a Gaussian $\mathcal{N}(0, 3)$ at each timestep (as we do in our experiments in Section \ref{sec:experiments}), and fix agent 1 to always truthfully signal the action it will take.  
In this case, the optimal policy for agent 2 is to select the best response to agent 1's action, which it knows exactly because agent 1 always signals truthfully. 
Depending on the payoff matrix $R$, sometimes the best response will be to take action 1, and sometimes to take action 2. 
Clearly agent 2's policy exhibits significant positive listening (it changes its action depending on the message from agent 1 and the input). However, if $R$ is drawn randomly, when averaged across inputs agent 2 will take action 1 and 2 equally. The IC will then be 0 in expectation, since it calculates $p(a,m)$ by averaging over episodes and does not condition on the input $R$.  






\vspace{-3mm}
\RestyleAlgo{algoruled}
\CommentSty{}
\begin{algorithm} 
\label{alg:cic}
\caption{One-step causal influence of communication. }
\KwData{Agent policy $\pol_1$, other agent policy $\pol_2$, possible messages $\bar{m}=(m_0, ..., m_{M-1})$, number of test games $T$.}
$\text{CIC} = 0$\\
\For{$i \in \{0, ...,  T - 1\}$} { 
    Generate new state $S$, observations $O$. \\
    \tcp{Intervene by changing message $m_j$}
    \For{$j \in \{0, ...,  M - 1\}$} { 
        $p(m_j) \leftarrow \pol_2(m_j|o_2)$, \hspace{3mm} $p(a | m_j) \leftarrow \pol_1(a | o_1, m_j)$\\
        $p(a, m_j) = p(a | m_j) p(m_j) $\\
        $p(a) = \sum_{m \in \mathcal{A}^m} p(a, m)$
        $\text{CIC} \mathrel{+}= 1/T \cdot \sum_{a \in \mathcal{A}^e} p(a, m_j) \log \frac{p(a, m_j)}{p(a)p(m_j)}$
    }
}
\end{algorithm}

\subsection{Causal influence of communication}
\label{sec:cic}

We now propose a metric that more directly measures positive listening, which
we call the causal influence of communication (CIC). This nomenclature follows concurrent work in \cite{jaques2018intrinsic}, where the authors also propose to train agents to maximize causal influence (both of actions and of messages), with good results on a number of domains. We recommend the reader consult \cite{jaques2018intrinsic} for a more thorough examination of causal influence. 

CIC measures the causal effect that one agent's message has on another agent's behaviour. In the simplest case, we can measure the effect that an agent's message has on the next action of the opponent. We call this one-step causal influence. The one-step CIC is calculated using the mutual information between an agent's message and the other agent's action, similarly as for IC (cf. Eq \ref{eq:sc}), however the probabilities $p(a, m) = \pol_1(a|m)\pol_2(m)$ represent the  
change in the agent's ($\pol_1$) action probability distribution when intervening to change the message $m$ spoken by the other agent ($\pol_2$). 
In other words, the probabilities are normalized over each game, rather than across games. 
We describe in detail how to calculate the one-step CIC in Algorithm 1, and we discuss how CIC might be generalized to the multi-step case in Section \ref{sec:general}.
When calculating the CIC, care must be taken that we condition on all the variables that can affect the other agent's action, to avoid back-door paths \cite{pearl2016causal}. In our setting this is easy, as our MCGs are not iterated (the state and reward are independent of actions and states at previous timesteps). 



\section{Experimental setup}

\subsection{Matrix Communication Games}
\label{sec:mcgs}

Our work is based on a simple class of games called \textit{matrix games}, where each agent $i$'s reward $r_i^t$ at time step $t$ is determined via lookup into a fixed payoff matrix $R_i^t \in \mathbb{R}^{|\mathcal{A}^e_1| \times \cdots \times  |\mathcal{A}^e_N|}$ for each agent, indexed by the agents' actions (i.e.\@ $r_i^t = R_i^t(a_1^t,... a_N^t)$). 
We study emergent communication in an augmented version of matrix games which we call \textit{matrix communication games} (MCGs), where agents can send discrete, costless messages over a communication channel before acting. In other words, MCGs are matrix games where $|\mathcal{A}^m_i| > 0$ for some $i \in \{0, ..., N\}$.  
MCGs are interesting to study because they are the simplest form of game where agents can both communicate with each other, and act in the environment. 
MCGs can be easily adapted to various settings (cooperative vs. non-cooperative, fully vs. partially observable), are simple and efficient for learning, and the resulting communication policies are generally interpretable. 
Variants of matrix games with communication have been used to model emergent signaling in the fields of economics, game theory, and evolutionary biology \cite{farrell1996cheap,wagner2003progress,smith1991honest,huttegger2010evolutionary}.

In our variant of the game, two agents play an MCG in rounds. A round begins with one agent communicating, and ends when both agents have acted and received a reward, at which point another payoff $R^{t+1} = (R_1^{t+1}, R_2^{t+1})$ is selected. 
The agents observe both payoff matrices and all messages sent that round. Communication occurs in turns;\footnote{The exact form of communication is not important for our results; we have observed similar behaviour when agents speak simultaneously, or if only one agent speaks.} at each round, one agent is randomly chosen to speak first, and sends a message (a one-hot vector of length $M=|\mathcal{A}^m|$) to the other agent. That agent observes the message from the first agent, and can in turn send its own message. After this exchange, both agents act simultaneously, and receive a reward. We consider the non-iterated case, where the actions and messages from previous rounds are not observed.\footnote{In Section \ref{sec:a2c} we show that the iterated case where agents (trained using the A2C algorithm \cite{mnih2016asynchronous}) have a memory of previous interactions produces similar behaviour.} To simplify learning, we use this reward to update both the agent's action and communication policies for that round. 

For the experiments in this paper, we consider the \textit{general-sum} case, where agents are not strictly cooperative and may have competing objectives (i.e.\@ $R_1 \neq R_2$). Note that this does not mean the agents are strictly competitive either; the agents have some \textit{partial common interest} in the game which permits a degree of coordination.  See the Appendix for examples of matrix games.
We vary the size of the payoff matrix from 2x2, to 4x4, to 8x8, and we provide a communication channel slightly larger than the number of actions ($M$ = 4, 6, and 10, respectively), so that the agents can in theory learn flexible communication policies. 

\subsection{Model and learning algorithm}
\label{sec:algo}

We train our agents using an adaptation of the REINFORCE algorithm \citep{williams1992simple}. We represent the agents' policies using a two-layer feed-forward network (with parameters $\theta^n$), with separate linear output layers for the action $a$ ($\theta^a$), communication $c$ ($\theta^c$), and learned baseline $V$ ($\theta^v$). Let $\theta^\mathbf{a}=\{\theta^a,\theta^n\}$ be all the parameters of the action network giving rise to an action policy $\pol_{\theta^\mathbf{a}}$, and similarly for $\theta^\mathbf{c}$ and $\theta^\mathbf{v}$ (with outputs $\pol_{\theta^\mathbf{c}}$ and $V$). The objective for each agent can be written as:
\begin{align*}
J(\theta) = &J_{pol}(\theta^\mathbf{a}) + \lambda_{c} J_{pol}(\theta^\mathbf{c}) + \lambda_{ent} (J_{ent}(\theta^\mathbf{a}) + J_{ent}(\theta^\mathbf{c})) + 
\lambda_v J_{v}(\theta^\mathbf{v}),& 
\end{align*}
where $\theta = \{\theta^n, \theta^a, \theta^c, \theta^v\}$. This can be broken down as follows: $J_{pol}(\theta^\mathbf{a}) = \mathbb{E}_{\pi_{\theta^\mathbf{a}}} [-\log \pi_{\theta^\mathbf{a}}(a|o) \cdot (r - V(o))]$ is the normal REINFORCE update for the action probabilities, with the learned value function $V(o), o \in \mathcal{O}$ as a baseline to reduce the variance. $J_{pol}(\theta^\mathbf{c})$ takes the same form, except $V$ only uses information available to the agent at the time it sends its message, which may not include the message of the other agent (there is no separate value function for the communication output).
This value function is updated using the squared loss $J_{v}(\theta^\mathbf{v}) = (r - V(o))^2$.
The entropy bonus $J_{ent}(\theta)$ gives a small reward based on the entropy of the agent's policies, as is common in the RL literature to encourage exploration \cite{williams1991function,schulman2015trust,christiano2017deep}. 
Each $\lambda_i$ is a real-valued coefficient that we tune separately. We provide full hyperparameter and training details in the Appendix.

\section{Results}
\label{sec:experiments}

We now train deep RL agents to play MCGs. We find that even though our agents show strong indicators of communicating according to speaker consistency and qualitative analysis, 
this does not mean that the communication is useful. In fact, we show that this `communication' occurs even if the messages are scrambled (replaced by a random message) before being observed. 




\begin{figure}
    \centering
    \includegraphics[width=0.45\textwidth]{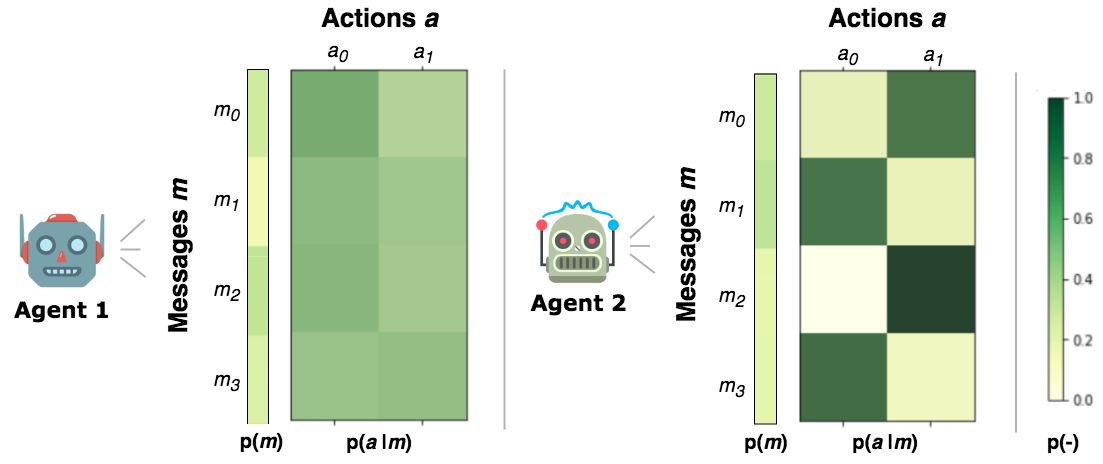}
    \vspace{-2mm}
    \caption{Visualization of example learned policies for two agents playing 2x2 MCGs ($M=4$), averaged over 1000 test games. Agent 2 learns to clearly signal its next action. \vspace{-3mm}} 
    \label{fig:vis}
\end{figure}

\subsection{Positive signaling with random payoffs}
\label{sec:5.1}

\paragraph{Fixed $R$ setting}
It is known that humans are able to use communication to obtain a higher reward in  general-sum matrix games, such as the Prisoner's Dilemma \cite{sally1995conversation}. To our knowledge, whether RL agents can learn to communicate for various MCGs remains an open question. 
So, we first conduct an experiment where we train two REINFORCE agents to play MCGs where the payoff $R^t$ is fixed for every timestep $t$. We vary the size of the payoff matrix and communication channel as described in Section \ref{sec:mcgs}. 
We find experimentally that, for every payoff matrix we tried, the agents don't learn to communicate. 
Instead, when there is partial common interest, agents collapse to executing a single action (even with a well-tuned $\lambda_{ent}$), and in zero-sum games they cycle between actions without communicating. 
Intuitively, this makes sense; the main utility in an agent learning to communicate in this setting is in reducing the other agent's uncertainty about their action. Evidently, when always playing the same payoff it is easier for these na{\"i}ve agents to adapt to the actions of the opponent directly, rather than learning a communication protocol.

\paragraph{Randomized $R$ setting}
One way we can increase the uncertainty each agent has about the other's action (with the hope of producing emergent communication) is by \textit{randomizing the payoffs} at each round. In our next experiment, we train two agents on an MCG where, at every round, each entry of the payoff matrix $R_i^t$ is drawn from a Gaussian distribution $\mathcal{N}(\mu, \sigma^2)$, with $\mu=0$ and $\sigma^2=3$.

As shown in Figure \ref{fig:bar_chart}, in this randomized setting speaker consistency emerges. The SC for the generated messages is significantly greater than for random messages, which rules out the possibility that the agents' messages are simply acting as a random public signal that the agents are using to condition their actions (as would be the case in a correlated equilibrium \cite{aumann1974subjectivity}). The CI is also higher for the generated messages, and the entropy of the message distribution is lower, both indicating that communication has emerged. We also examine the policies qualitatively in Figure \ref{fig:vis}, and find that one agent clearly learns to signal its intended action. 
One plausible explanation for this phenomenon is that the agents have uncertainty about the intended action of the other agent, and receiving a high reward in these games requires agents to predict the intended action of the opponent. 
An agent can do this in two ways: first by using the opponent's payoff matrix (and its own) to determine what action they are likely to take to achieve a high reward, and second using the opponent's message to predict their most likely action. When the payoff matrix is changing at each time step, predicting the action of the opponent conditioned solely on the payoff matrix becomes significantly harder. Thus, the agents seem to learn to leverage the communication channel to provide additional information to the other agent, which is beneficial for both agents when the game has partial common interest.


\begin{table} 
    \centering
    \begin{tabular}{c c c c } \toprule
        \textbf{Payoff} & $\min(\textbf{CIC})$ & \textbf{Average} & \textbf{\% games with} \\
        \textbf{Size} &  & \textbf{CIC} &  \textbf{CIC < $1.02 \cdot \min(\text{CIC})$} \\ \hline
        2x2 & 1.386 & 1.408 $\pm$ 0.002 & 89.3 $\pm$ 0.6\% \\
        4x4 & 1.792 & 1.797 $\pm$ 0.001 & 97.9 $\pm$ 0.4\% \\
        8x8 & 2.303 & 2.303 $\pm$ 0.001 & 99.9 $\pm$ 0.1\% \\ \bottomrule
    \end{tabular}
    \caption{Causal influence values for various matrix sizes, calculated over 1000 test games. In all cases, the average CIC is very close to the minimum CIC (when changing the message has no effect on the action distribution).\vspace{-3mm}}
    \label{tab:cic}
\end{table}

\begin{table}[]
    \centering
    \begin{tabular}{c c c c} \toprule
    \textbf{Experiments} & \textbf{2x2 payoff} & \textbf{4x4 payoff} & \textbf{8x8 payoff} \\ \hline
        
       Scrambled $c$ &   0.198 $\pm$ 0.038  &  0.487 $\pm$ 0.051  & 0.597 $\pm$ 0.091  \\
       Separate $c$ net &   0.028 $\pm$ 0.002  & 0.124 $\pm$ 0.011  & 0.020 $\pm$ 0.019  \\
        No $c$ training&  0.171 $\pm$ 0.033  & 0.428 $\pm$ 0.025 & 0.686 $\pm$ 0.049  \\ \hline
        Default & 0.202 $\pm$ 0.040  &  0.510 $\pm$ 0.094  &  0.541 $\pm$ 0.090  \\
        \bottomrule
    \end{tabular}
    \caption{SC values for the randomized $R$ setting. `Scrambled $c$' is  when the messages are replaced by a random message before being observed, `Separate $c$ net' is when the action and message networks have no shared parameters, and `No $c$ training' is when $\lambda_{c}=0$. \vspace{-3mm}}
    \label{tab:results}
\end{table}

\subsection{Positive signaling $\centernot\implies$ positive listening}
It turns out that, in the randomized $R$ setting, the communication has very little effect on the agents' behaviour. We show this in two ways.
First, we examine the trained policies directly to see how altering the message in various games changes the resulting action selection, using the CIC metric. 
We calculate the CIC over 1000 test games, and show the results in Table \ref{tab:cic}. We find that, for the vast majority of games, the message sent by an agent has no effect on the action of the opponent. Thus, communication is not having a significant effect on the training process. 

\begin{figure}[]
    \centering
    \vspace{-3mm}
    \includegraphics[width=0.45\textwidth]{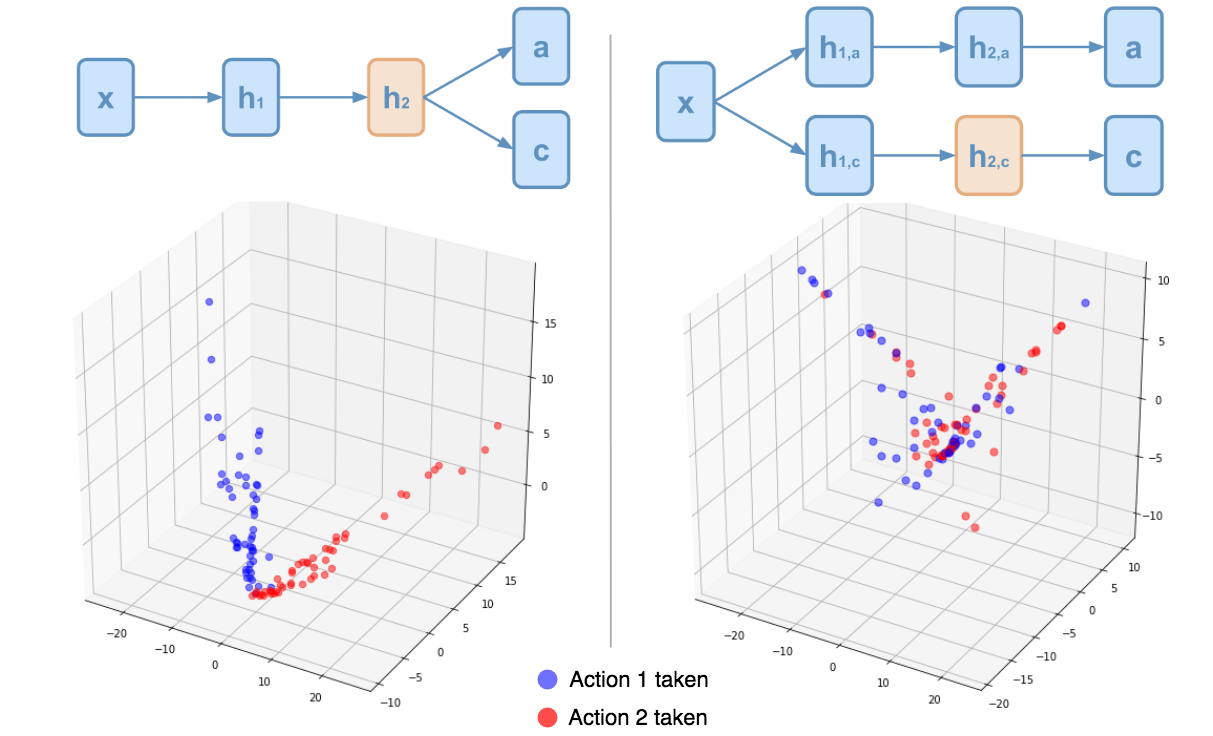}
    \caption{Activations of the last layer of the policy network for both the standard architecture (left), and when using a separate network for communication (right). Calculated on 100 random 2x2 games, and reduced to 3D using PCA. }
    \label{fig:hscatter}
\end{figure}

Second, we conduct an experiment where we train the agents in the randomized $R$ setting described above, except we \textit{scramble} the messages received by both agents. That is, each agent outputs some message $c_i$, but we have both agents observe a different communication $c'_i$, which has no relation to $c_i$ and is drawn uniformly at random from the set of messages. Thus, there is no possibility for the agents' messages to impact the learning whatsoever. However, as shown in Table \ref{tab:results}, the SC between the agents' action and sent message (\textit{not} the random replacement message) is still positive and indistinguishable from the SC in the regular MCG set-up. This is convincing evidence that the correlation between actions and communications does not emerge because the message are useful, but rather as a byproduct of optimization. 

\begin{figure*}[]
    \centering
    \includegraphics[width=0.32\textwidth]{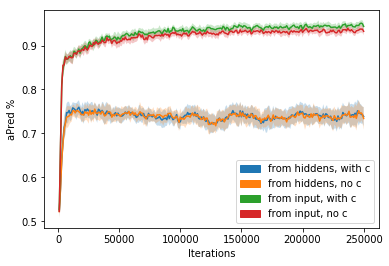}
    \includegraphics[width=0.32\textwidth]{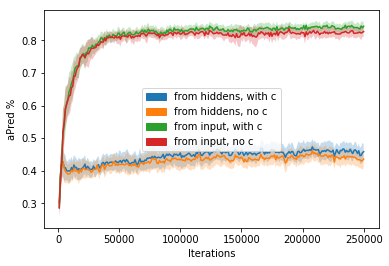}
    \includegraphics[width=0.32\textwidth]{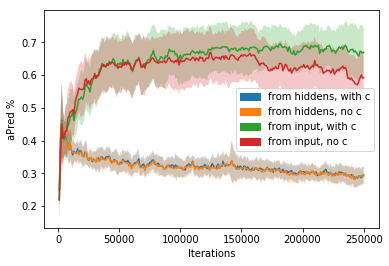}
    \vspace{-2mm}
    \caption{Accuracy in predicting the opponent's action using action classifier probes, for 2x2 payoffs (left), 4x4 payoffs (center), and 8x8 payoffs (right). `no c' indicates that the other agent's communication was not used to predict their action, and `from input' indicates that a separate network was trained to predict the action (rather than using a linear model on top of the last hidden layer of the policy network).}
    \label{fig:apreds}
\end{figure*}


\paragraph{Why is the SC positive?} 
If the emergent communication is not useful at all, why is the SC positive? 
To help us understand what the policies are learning, we train agents according to the randomized $R$ setting in Section \ref{sec:5.1}, and 
we plot the activations of the last hidden layer (values in the policy network after the last activation function) for 100 inputs using principal component analysis (PCA) \cite{pearson1901principal} in Figure \ref{fig:hscatter}.    
This shows us our policy network's learned representations of the data \cite{lecun2015deep}.  
When using shared features (left), the network learns to separate the inputs based on which action the agent takes. This makes sense: in order to take the appropriate action for a given input, the representations need to be linearly separable at the last layer of the network, since the action output layer has a linear decision boundary \cite{goodfellow2014generative}. This separation of representations does not occur in the last layer of a separate communication network, which provides further evidence for this explanation. 




This separation of representations makes it easy for a relationship to emerge between an agent's actions and messages, even if the parameters of the communication head are completely random; since the communication output layer also has a linear decision boundary, it is likely to separate based on intended actions to some degree. Indeed, we find that SC emerges using our architecture when the communication parameters are not trained (Table \ref{tab:results}).
Further, when we re-train the agents using completely separate networks for the actions and messages, we find that the SC completely disappears (see Table \ref{tab:results}), showing that it was indeed our choice of architecture that resulted in the emergent signaling behaviour.

\paragraph{Why aren't the agents using the messages?}
We would imagine that, even if the correlation between actions and messages emerged accidentally, that this might still be useful for the agents in selecting their action. After all, isn't more information about the opponent strictly better? To answer this question, we use a set of \textit{action classifier probes}, related to the linear classifier probes for understanding hidden layers of neural networks proposed in \cite{alain2016understanding}. Specifically, in the randomized $R$ setting, we train a neural network `probe' to predict the action of the opponent in the current round, based on either part of the input (using 2-layer MLP probes) or part of the last hidden layer of the policy network (using linear probes). 

The results are shown in Figure \ref{fig:apreds}. We observe that removing the opponent's message from the input does not significantly reduce the accuracy in predicting the opponent's action, both when using a probe directly from the input and a probe from the last layer of the network. This suggests that the information being provided by the messages is redundant when compared to the information provided by the payoff matrix itself.

\begin{table}[]
    \centering
    \begin{tabular}{c c c c} \toprule
    \textbf{Experiments} & \textbf{2x2 payoff} & \textbf{4x4 payoff} & \textbf{8x8 payoff} \\ \hline
        
       Scrambled $c$ &   0.195 $\pm$ 0.059  &  0.127 $\pm$ 0.065  & 0.208 $\pm$ 0.067  \\
       Separate $c$ net &   0.016$\pm$ 0.019  & 0.019 $\pm$ 0.011  & 0.000 $\pm$ 0.000  \\
         \hline
        Default & 0.169 $\pm$ 0.015  &  0.130 $\pm$ 0.055  &  0.269 $\pm$ 0.076  \\
        \bottomrule
    \end{tabular}
    \caption{Speaker consistency values for different experiments in the iterated MCG case with A2C agents \cite{mnih2016asynchronous}. \vspace{-7mm}}
    \label{tab:a2cresults}
\end{table}

\subsection{Results in the iterated MCG setting}
\label{sec:a2c}
The results in our paper are not limited to the non-iterated case. To show this, we run experiments on an iterated version of our environment, using the A2C algorithm \cite{mnih2016asynchronous}. We keep the policy architectures the same, except we give each agent a memory of the previous 5 rounds (the actions and the messages of both agents), which is concatenated to the input at each round. We increase the  discount factor $\gamma$ to 0.9. Changing our REINFORCE algorithm (Eq. \ref{eq:reinforce}) to A2C requires changing the way $Q^{\pol}$ is estimated; instead of using the next reward, we use the $n$-step return (sum of $n$ next rewards), with $n=5$ \cite{mnih2016asynchronous}. The results are shown in Table \ref{tab:a2cresults}. We can see that the same general trend is present: there is positive SC in the randomized $R$ setting, even when scrambling the messages $c$; however, the SC disappears when a different network is used to produce the messages.

\begin{table*}[]
\small
    \centering
    \begin{tabular}{c c c c c l} \toprule
          &  \multicolumn{2}{c}{\textbf{Positive signaling}} & \multicolumn{2}{c}{\textbf{Positive listening}} &   \\
        \textbf{Metric} & Sufficient? & Necessary? & Sufficient? & Necessary? & \textbf{Remarks} \\ \hline 
        SC & \cmark & \xmark & \xmark$^*$ & \xmark  & Not necessary for positive signaling, as there may be no relationship between the\\
        &  & & & & message and subsequent action (can communicate previous observations or actions). \\
        CI & \cmark & \xmark & \xmark$^*$ & \xmark & More restrictive than SC, punishes formation of synonyms. \\
        $H(\pol^c)$  & \xmark & \xmark & \xmark$^*$ & \xmark  & Useful for diagnosing policy behaviour, but not as a metric. \\
        IC & \xmark & \xmark & \cmark & \xmark$^*$  & Since it is not state-dependent, IC can miss many positive listening relationships. \\
        MIN & \xmark & \xmark & \xmark$^*$ & \cmark & If $||W^1_m||=0$, then no positive listening is present.\\  
        $\Delta r$ & \cmark & \xmark & \cmark & \xmark  & Should always be measured, strong indicator that communication is present. But less \\
         &  & & & & applicable when comparing the communicative behaviour of two policies \\
        Qual. & \cmark & \cmark & \xmark$^*$ & \xmark   & Very useful for understanding agent behaviour. Can come in many forms.  \\
        CIC & \xmark & \xmark$^*$ & \cmark & \cmark   & Should always be used to determine effect communication has on agent behaviour. \\
    \bottomrule
    \end{tabular}
    \caption{Summary of the metrics analyzed in this paper ($\Delta r$ = increase in reward when adding a communication channel, Qual. = qualitative analysis of messages, and MIN = message input norm, detailed in Section \ref{sec:recs}). Asterisks ($^*$) mark relationships we have shown experimentally (Sections \ref{sec:experiments} and \ref{sec:recs}) or via counterexample (Section \ref{sec:metrics}).  See text for a more detailed explanation. \vspace{-7mm}}
    \label{tab:discussion}
\end{table*}

\section{Discussion}

\subsection{How general is this analysis?}
\label{sec:general}

\paragraph{Positive signaling without positive listening}
An important question is whether the behaviours observed in this paper are specific to training our policy architecture on MCGs, and whether any of these insights can be applied in other emergent communication settings. We conjecture that this could indeed happen whenever the agent's architecture uses shared feature learning layers between the action and communication outputs; policies will always learn representations that separate the inputs based on the desired action, and this may lead to spurious correlations between an agent's messages and actions. Since sharing features is quite common in RL (e.g.\@ \cite{jaderberg2016reinforcement}), it is possible that this becomes an occasional occurrence in emergent communication work. 

However, our claim is not that this specific failure case will be frequently observed; rather, our goal is to highlight the importance of understanding what our metrics are measuring, and encourage emergent communication researchers to explore quantitative metrics for measuring the impact that the agents' communication is having on their behaviour (i.e.\@ positive listening).  

\paragraph{Scaling causal influence}
In this paper, we focused on the one-step approximation to CIC, which only calculates the effect of an agent's message (which consists of one symbol) on the other agent's next action. While this is sufficient for the non-iterated MCG setting, as we move to more complex environments we will need to measure the effect of compositional messages on the long-term behaviour of all other agents in the environment. In this case, calculating CIC na{\"i}vely using Algorithm 1 will be computationally expensive.
However, we can make this efficient via \textit{sampling}; rather than iterating over all possible messages and all agents, the CIC could be calculated by iterating over a small set of messages sampled from the agent's communication policy, and evaluating the change in behaviour over finite time horizon for agents in some neighbourhood of the speaking agent. We leave a detailed exploration of this direction to future work. 







\subsection{Recommendations} 
\label{sec:recs}


Here we provide recommendations as to when the metrics 
presented in Section \ref{sec:metrics} might be used to either detect whether communication is emerging, or measure the difference in communication quality between algorithms. We also propose some other tests that could be used for this purpose. 
We summarize our insights in Table \ref{tab:discussion}. In general, no single metric or test will tell the whole story, and we advise using several of them to illuminate the agents' behaviour. 


\paragraph{Detecting emergent communication}
If our goal is to detect whether communication is emerging at all, showing that adding a communication channel to a given algorithm leads to improved reward is a sufficient indicator. However, it may not be necessary; agents may obtain a similar reward by coordinating via learned convention \cite{lerer2018learning}, rather than communication. In other words, communication may act as an alternate pathway for optimization.  
Detecting useful communication is not as simple as testing if removing the communication channel at test time leads to a decrease in reward; neural networks are notoriously sensitive to their input distribution \cite{szegedy2013intriguing}, and a change in this distribution (e.g.\@ setting the messages to 0) may cause them to fail, even if the messages contain no useful information. We recommend instead using CIC and other causal metrics, discussed below. 


The variant of SC explored here only measures the one-step relationship between an agent's message and subsequent action. In general, communication could influence the actions of agents further in the future than a single time step, and the language used by the agents may be compositional and temporally extended. This should be taken into account as we move towards more complex environments. In general, we acknowledge that speaker consistency is useful from the perspective of detecting positive signaling, but we reiterate that the observed relationships may be spurious. 


There are other aspects of the environment that an agent could learn to signal about: an agent might send a message to get another agent to perform an action, to share an observation it has made, or to reveal the sequence of actions it has taken in the past. New metrics need to be developed to evaluate these possibilities. We recommend researchers evaluate the quantities that are relevant for their environment or task. 
The crucial point is that, if these quantities are measured by observing the agent's behaviour without causal intervention, detecting positive listening is difficult, as evidenced by our experiments in Section \ref{sec:experiments}. If a relationship is observed between an agent's messages and some quantity in the environment, we recommend researchers \textit{investigate the causal relationship} between these variables, by intervening to change in turn the environmental quantity and the agent's message, and observing the impact on the other quantity over a number of episodes \cite{everitt2019understanding}. 

One way to tell conclusively that there is no positive listening is to look at the weight matrix of the first layer of the policy network ($W^1$), specifically the part that comes from the message inputs of the other agents ($W^1_m$). If the norm of this part of the weight matrix (the \textit{message input norm}, MIN) is 0, then clearly no positive listening is present, as the messages from the other agents cannot affect a given agent's behaviour. However, just because $||W^1_m||>0$, does not mean positive listening is present; in our experiments on MCGs, we found that this norm was of similar magnitude to the norm of the weights from the payoff matrix.

\paragraph{Measuring improvement in communication}

How should we judge the quality of a learned communication protocol in a multi-agent environment? Of course, this depends on the environment and the objectives of the researcher. Often, researchers may want to show that their algorithm exhibits a new kind of communication (e.g. verbal agreements, compositional language, or deception). In these cases, it makes sense to use metrics targeted at measuring the phenomenon in question. If the goal is to develop compositional communication, as has been the case for several recent emergent communication papers \cite{mordatch2017emergence,havrylov2017emergence,bogin2018emergence}, then it is perhaps sufficient to evaluate using metrics designed to measure compositionality, such as context independence \cite{bogin2018emergence}. These metrics will have to be developed on a case-by-case basis, depending on the type of communication under investigation. 

There may also be cases where we simply want to show that the learned communication protocol for a proposed algorithm has a larger effect on agent behaviour than for previous algorithms. Here, a variant of CIC should be used that measures the impact of communication on the long-term behaviour of the other agents. 
Another test one could run is measuring the difference in reward for each algorithm with and without communication. 
This should be done by training each algorithm from scratch with and without communication, rather than removing the communication channel at test time, to avoid the problems of distributional shift mentioned earlier in this section. 
Of course, since these metrics may be exploitable, it is important to benchmark against a range of metrics and tests to avoid overfitting.


\section{Conclusion}
In this paper, we have highlighted some of the challenges of measuring emergent communication, namely that existing metrics for detecting positive signaling do not necessarily indicate the presence of positive listening. 
In addition to measuring the reward and studying agent behaviour qualitatively, we advocate for investigating the causal relationship between the agents' messages and other variables in the environment by performing interventions. This gives rise to a family of metrics, including the CIC described here, which evaluates the impact of an agent's message on another agent's next action. An open question is how our insights can be applied to evaluating agent communication with humans. Indeed, while there have been some methods proposed for understanding agent communication by `translating' it to human language \cite{andreas2017translating}, this remains an open problem, and we leave an examination of this direction to future work.

\section*{Acknowledgements}
The authors would like to thank Ethan Perez, Ari Morcos, Alex Peysakhovitch, Jean Harb, Evgeny Naumov, Michael Noukhovitch, Max Smith, and other researchers at FAIR and MILA for discussions related to the paper and comments on early drafts. We thank Natasha Jaques for answering our questions regarding the calculation of instantaneous coordination. We also thank the Pytorch development team \citep{paszke2017automatic}. R.L. is supported by a Vanier Scholarship.

\bibliographystyle{ACM-Reference-Format}  
\bibliography{aamas19}  


\begin{thebibliography}{00}


\ifx \showCODEN    \undefined \def \showCODEN     #1{\unskip}     \fi
\ifx \showDOI      \undefined \def \showDOI       #1{#1}\fi
\ifx \showISBNx    \undefined \def \showISBNx     #1{\unskip}     \fi
\ifx \showISBNxiii \undefined \def \showISBNxiii  #1{\unskip}     \fi
\ifx \showISSN     \undefined \def \showISSN      #1{\unskip}     \fi
\ifx \showLCCN     \undefined \def \showLCCN      #1{\unskip}     \fi
\ifx \shownote     \undefined \def \shownote      #1{#1}          \fi
\ifx \showarticletitle \undefined \def \showarticletitle #1{#1}   \fi
\ifx \showURL      \undefined \def \showURL       {\relax}        \fi
\providecommand\bibfield[2]{#2}
\providecommand\bibinfo[2]{#2}
\providecommand\natexlab[1]{#1}
\providecommand\showeprint[2][]{arXiv:#2}

\bibitem[\protect\citeauthoryear{Alain and Bengio}{Alain and Bengio}{2016}]%
        {alain2016understanding}
\bibfield{author}{\bibinfo{person}{Guillaume Alain} {and}
  \bibinfo{person}{Yoshua Bengio}.} \bibinfo{year}{2016}\natexlab{}.
\newblock \showarticletitle{Understanding intermediate layers using linear
  classifier probes}.
\newblock \bibinfo{journal}{{\em arXiv preprint arXiv:1610.01644\/}}
  (\bibinfo{year}{2016}).
\newblock


\bibitem[\protect\citeauthoryear{Andreas, Dragan, and Klein}{Andreas
  et~al\mbox{.}}{2017}]%
        {andreas2017translating}
\bibfield{author}{\bibinfo{person}{Jacob Andreas}, \bibinfo{person}{Anca
  Dragan}, {and} \bibinfo{person}{Dan Klein}.} \bibinfo{year}{2017}\natexlab{}.
\newblock \showarticletitle{Translating neuralese}.
\newblock \bibinfo{journal}{{\em arXiv preprint arXiv:1704.06960\/}}
  (\bibinfo{year}{2017}).
\newblock


\bibitem[\protect\citeauthoryear{Aumann}{Aumann}{1974}]%
        {aumann1974subjectivity}
\bibfield{author}{\bibinfo{person}{Robert~J Aumann}.}
  \bibinfo{year}{1974}\natexlab{}.
\newblock \showarticletitle{Subjectivity and correlation in randomized
  strategies}.
\newblock \bibinfo{journal}{{\em Journal of mathematical Economics\/}}
  \bibinfo{volume}{1}, \bibinfo{number}{1} (\bibinfo{year}{1974}),
  \bibinfo{pages}{67--96}.
\newblock


\bibitem[\protect\citeauthoryear{Bogin, Geva, and Berant}{Bogin
  et~al\mbox{.}}{2018}]%
        {bogin2018emergence}
\bibfield{author}{\bibinfo{person}{Ben Bogin}, \bibinfo{person}{Mor Geva},
  {and} \bibinfo{person}{Jonathan Berant}.} \bibinfo{year}{2018}\natexlab{}.
\newblock \showarticletitle{Emergence of communication in an interactive world
  with consistent speakers}.
\newblock \bibinfo{journal}{{\em arXiv preprint arXiv:1809.00549\/}}
  (\bibinfo{year}{2018}).
\newblock


\bibitem[\protect\citeauthoryear{Brown, Pietra, Pietra, and Mercer}{Brown
  et~al\mbox{.}}{1993}]%
        {brown1993mathematics}
\bibfield{author}{\bibinfo{person}{Peter~F Brown}, \bibinfo{person}{Vincent
  J~Della Pietra}, \bibinfo{person}{Stephen A~Della Pietra}, {and}
  \bibinfo{person}{Robert~L Mercer}.} \bibinfo{year}{1993}\natexlab{}.
\newblock \showarticletitle{The mathematics of statistical machine translation:
  Parameter estimation}.
\newblock \bibinfo{journal}{{\em Computational linguistics\/}}
  \bibinfo{volume}{19}, \bibinfo{number}{2} (\bibinfo{year}{1993}),
  \bibinfo{pages}{263--311}.
\newblock


\bibitem[\protect\citeauthoryear{Cao, Lazaridou, Lanctot, Leibo, Tuyls, and
  Clark}{Cao et~al\mbox{.}}{2018}]%
        {cao2018emergent}
\bibfield{author}{\bibinfo{person}{Kris Cao}, \bibinfo{person}{Angeliki
  Lazaridou}, \bibinfo{person}{Marc Lanctot}, \bibinfo{person}{Joel~Z Leibo},
  \bibinfo{person}{Karl Tuyls}, {and} \bibinfo{person}{Stephen Clark}.}
  \bibinfo{year}{2018}\natexlab{}.
\newblock \showarticletitle{Emergent Communication through Negotiation}.
\newblock \bibinfo{journal}{{\em arXiv preprint arXiv:1804.03980\/}}
  (\bibinfo{year}{2018}).
\newblock


\bibitem[\protect\citeauthoryear{Choi, Lazaridou, and de~Freitas}{Choi
  et~al\mbox{.}}{2018}]%
        {choi2018compositional}
\bibfield{author}{\bibinfo{person}{Edward Choi}, \bibinfo{person}{Angeliki
  Lazaridou}, {and} \bibinfo{person}{Nando de Freitas}.}
  \bibinfo{year}{2018}\natexlab{}.
\newblock \showarticletitle{Compositional Obverter Communication Learning From
  Raw Visual Input}.
\newblock \bibinfo{journal}{{\em arXiv preprint arXiv:1804.02341\/}}
  (\bibinfo{year}{2018}).
\newblock


\bibitem[\protect\citeauthoryear{Christiano, Leike, Brown, Martic, Legg, and
  Amodei}{Christiano et~al\mbox{.}}{2017}]%
        {christiano2017deep}
\bibfield{author}{\bibinfo{person}{Paul~F Christiano}, \bibinfo{person}{Jan
  Leike}, \bibinfo{person}{Tom Brown}, \bibinfo{person}{Miljan Martic},
  \bibinfo{person}{Shane Legg}, {and} \bibinfo{person}{Dario Amodei}.}
  \bibinfo{year}{2017}\natexlab{}.
\newblock \showarticletitle{Deep reinforcement learning from human
  preferences}. In \bibinfo{booktitle}{{\em Advances in Neural Information
  Processing Systems}}. \bibinfo{pages}{4299--4307}.
\newblock


\bibitem[\protect\citeauthoryear{Das, Kottur, Moura, Lee, and Batra}{Das
  et~al\mbox{.}}{2017}]%
        {das2017learning}
\bibfield{author}{\bibinfo{person}{Abhishek Das}, \bibinfo{person}{Satwik
  Kottur}, \bibinfo{person}{Jos{\'e}~MF Moura}, \bibinfo{person}{Stefan Lee},
  {and} \bibinfo{person}{Dhruv Batra}.} \bibinfo{year}{2017}\natexlab{}.
\newblock \showarticletitle{Learning cooperative visual dialog agents with deep
  reinforcement learning}.
\newblock \bibinfo{journal}{{\em arXiv preprint arXiv:1703.06585\/}}
  (\bibinfo{year}{2017}).
\newblock


\bibitem[\protect\citeauthoryear{Everitt, Ortega, Barnes, and Legg}{Everitt
  et~al\mbox{.}}{2019}]%
        {everitt2019understanding}
\bibfield{author}{\bibinfo{person}{Tom Everitt}, \bibinfo{person}{Pedro~A
  Ortega}, \bibinfo{person}{Elizabeth Barnes}, {and} \bibinfo{person}{Shane
  Legg}.} \bibinfo{year}{2019}\natexlab{}.
\newblock \showarticletitle{Understanding Agent Incentives using Causal
  Influence Diagrams, Part I: Single Action Settings}.
\newblock \bibinfo{journal}{{\em arXiv preprint arXiv:1902.09980\/}}
  (\bibinfo{year}{2019}).
\newblock


\bibitem[\protect\citeauthoryear{Evtimova, Drozdov, Kiela, and Cho}{Evtimova
  et~al\mbox{.}}{2017}]%
        {evtimova2017emergent}
\bibfield{author}{\bibinfo{person}{Katrina Evtimova}, \bibinfo{person}{Andrew
  Drozdov}, \bibinfo{person}{Douwe Kiela}, {and} \bibinfo{person}{Kyunghyun
  Cho}.} \bibinfo{year}{2017}\natexlab{}.
\newblock \showarticletitle{Emergent Communication in a Multi-Modal, Multi-Step
  Referential Game}.
\newblock \bibinfo{journal}{{\em arXiv preprint arXiv:1705.10369\/}}
  (\bibinfo{year}{2017}).
\newblock


\bibitem[\protect\citeauthoryear{Farrell and Rabin}{Farrell and Rabin}{1996}]%
        {farrell1996cheap}
\bibfield{author}{\bibinfo{person}{Joseph Farrell} {and}
  \bibinfo{person}{Matthew Rabin}.} \bibinfo{year}{1996}\natexlab{}.
\newblock \showarticletitle{Cheap talk}.
\newblock \bibinfo{journal}{{\em Journal of Economic perspectives\/}}
  \bibinfo{volume}{10}, \bibinfo{number}{3} (\bibinfo{year}{1996}),
  \bibinfo{pages}{103--118}.
\newblock


\bibitem[\protect\citeauthoryear{Foerster, Song, Hughes, Burch, Dunning,
  Whiteson, Botvinick, and Bowling}{Foerster et~al\mbox{.}}{2018}]%
        {foerster2018bayesian}
\bibfield{author}{\bibinfo{person}{Jakob Foerster}, \bibinfo{person}{Francis
  Song}, \bibinfo{person}{Edward Hughes}, \bibinfo{person}{Neil Burch},
  \bibinfo{person}{Iain Dunning}, \bibinfo{person}{Shimon Whiteson},
  \bibinfo{person}{Matthew Botvinick}, {and} \bibinfo{person}{Michael
  Bowling}.} \bibinfo{year}{2018}\natexlab{}.
\newblock \showarticletitle{Bayesian action decoding for deep multi-agent
  reinforcement learning}.
\newblock \bibinfo{journal}{{\em arXiv preprint arXiv:1811.01458\/}}
  (\bibinfo{year}{2018}).
\newblock


\bibitem[\protect\citeauthoryear{Foerster, Assael1, de~Freitas, and
  Whiteson}{Foerster et~al\mbox{.}}{2016}]%
        {foerster16a}
\bibfield{author}{\bibinfo{person}{Jakob~N. Foerster},
  \bibinfo{person}{Yannis~M. Assael1}, \bibinfo{person}{Nando de Freitas},
  {and} \bibinfo{person}{Shimon Whiteson}.} \bibinfo{year}{2016}\natexlab{}.
\newblock \showarticletitle{{Learning to Communicate to Solve Riddles with Deep
  Distributed Recurrent Q-Networks}}.
\newblock  (\bibinfo{year}{2016}).
\newblock


\bibitem[\protect\citeauthoryear{Gauthier and Mordatch}{Gauthier and
  Mordatch}{2016}]%
        {gauthier2016paradigm}
\bibfield{author}{\bibinfo{person}{Jon Gauthier} {and} \bibinfo{person}{Igor
  Mordatch}.} \bibinfo{year}{2016}\natexlab{}.
\newblock \showarticletitle{A paradigm for situated and goal-driven language
  learning}.
\newblock \bibinfo{journal}{{\em arXiv preprint arXiv:1610.03585\/}}
  (\bibinfo{year}{2016}).
\newblock


\bibitem[\protect\citeauthoryear{Gintis, Smith, and Bowles}{Gintis
  et~al\mbox{.}}{2001}]%
        {gintis2001costly}
\bibfield{author}{\bibinfo{person}{Herbert Gintis}, \bibinfo{person}{Eric~Alden
  Smith}, {and} \bibinfo{person}{Samuel Bowles}.}
  \bibinfo{year}{2001}\natexlab{}.
\newblock \showarticletitle{Costly signaling and cooperation}.
\newblock \bibinfo{journal}{{\em Journal of theoretical biology\/}}
  \bibinfo{volume}{213}, \bibinfo{number}{1} (\bibinfo{year}{2001}),
  \bibinfo{pages}{103--119}.
\newblock


\bibitem[\protect\citeauthoryear{Goodfellow, Pouget-Abadie, Mirza, Xu,
  Warde-Farley, Ozair, Courville, and Bengio}{Goodfellow et~al\mbox{.}}{2014}]%
        {goodfellow2014generative}
\bibfield{author}{\bibinfo{person}{Ian Goodfellow}, \bibinfo{person}{Jean
  Pouget-Abadie}, \bibinfo{person}{Mehdi Mirza}, \bibinfo{person}{Bing Xu},
  \bibinfo{person}{David Warde-Farley}, \bibinfo{person}{Sherjil Ozair},
  \bibinfo{person}{Aaron Courville}, {and} \bibinfo{person}{Yoshua Bengio}.}
  \bibinfo{year}{2014}\natexlab{}.
\newblock \showarticletitle{Generative adversarial nets}. In
  \bibinfo{booktitle}{{\em Advances in neural information processing systems}}.
\newblock


\bibitem[\protect\citeauthoryear{Havrylov and Titov}{Havrylov and
  Titov}{2017}]%
        {havrylov2017emergence}
\bibfield{author}{\bibinfo{person}{Serhii Havrylov} {and} \bibinfo{person}{Ivan
  Titov}.} \bibinfo{year}{2017}\natexlab{}.
\newblock \showarticletitle{Emergence of language with multi-agent games:
  learning to communicate with sequences of symbols}. In
  \bibinfo{booktitle}{{\em Advances in Neural Information Processing Systems}}.
  \bibinfo{pages}{2149--2159}.
\newblock


\bibitem[\protect\citeauthoryear{Huttegger, Skyrms, Smead, and
  Zollman}{Huttegger et~al\mbox{.}}{2010}]%
        {huttegger2010evolutionary}
\bibfield{author}{\bibinfo{person}{Simon~M Huttegger}, \bibinfo{person}{Brian
  Skyrms}, \bibinfo{person}{Rory Smead}, {and} \bibinfo{person}{Kevin~JS
  Zollman}.} \bibinfo{year}{2010}\natexlab{}.
\newblock \showarticletitle{Evolutionary dynamics of Lewis signaling games:
  signaling systems vs. partial pooling}.
\newblock \bibinfo{journal}{{\em Synthese\/}} \bibinfo{volume}{172},
  \bibinfo{number}{1} (\bibinfo{year}{2010}), \bibinfo{pages}{177}.
\newblock


\bibitem[\protect\citeauthoryear{Jaderberg, Mnih, Czarnecki, Schaul, Leibo,
  Silver, and Kavukcuoglu}{Jaderberg et~al\mbox{.}}{2016}]%
        {jaderberg2016reinforcement}
\bibfield{author}{\bibinfo{person}{Max Jaderberg}, \bibinfo{person}{Volodymyr
  Mnih}, \bibinfo{person}{Wojciech~Marian Czarnecki}, \bibinfo{person}{Tom
  Schaul}, \bibinfo{person}{Joel~Z Leibo}, \bibinfo{person}{David Silver},
  {and} \bibinfo{person}{Koray Kavukcuoglu}.} \bibinfo{year}{2016}\natexlab{}.
\newblock \showarticletitle{Reinforcement learning with unsupervised auxiliary
  tasks}.
\newblock \bibinfo{journal}{{\em arXiv preprint arXiv:1611.05397\/}}
  (\bibinfo{year}{2016}).
\newblock


\bibitem[\protect\citeauthoryear{Jaques, Lazaridou, Hughes, Gulcehre, Ortega,
  Strouse, Leibo, and de~Freitas}{Jaques et~al\mbox{.}}{2018}]%
        {jaques2018intrinsic}
\bibfield{author}{\bibinfo{person}{Natasha Jaques}, \bibinfo{person}{Angeliki
  Lazaridou}, \bibinfo{person}{Edward Hughes}, \bibinfo{person}{Caglar
  Gulcehre}, \bibinfo{person}{Pedro~A Ortega}, \bibinfo{person}{DJ Strouse},
  \bibinfo{person}{Joel~Z Leibo}, {and} \bibinfo{person}{Nando de Freitas}.}
  \bibinfo{year}{2018}\natexlab{}.
\newblock \showarticletitle{Intrinsic Social Motivation via Causal Influence in
  Multi-Agent RL}.
\newblock \bibinfo{journal}{{\em arXiv preprint arXiv:1810.08647\/}}
  (\bibinfo{year}{2018}).
\newblock


\bibitem[\protect\citeauthoryear{Kingma and Ba}{Kingma and Ba}{2014}]%
        {kingma2014adam}
\bibfield{author}{\bibinfo{person}{Diederik~P Kingma} {and}
  \bibinfo{person}{Jimmy Ba}.} \bibinfo{year}{2014}\natexlab{}.
\newblock \showarticletitle{Adam: A method for stochastic optimization}.
\newblock \bibinfo{journal}{{\em arXiv preprint arXiv:1412.6980\/}}
  (\bibinfo{year}{2014}).
\newblock


\bibitem[\protect\citeauthoryear{Kottur, Moura, Lee, and Batra}{Kottur
  et~al\mbox{.}}{2017}]%
        {kottur2017natural}
\bibfield{author}{\bibinfo{person}{Satwik Kottur}, \bibinfo{person}{Jos{\'e}~MF
  Moura}, \bibinfo{person}{Stefan Lee}, {and} \bibinfo{person}{Dhruv Batra}.}
  \bibinfo{year}{2017}\natexlab{}.
\newblock \showarticletitle{Natural Language Does Not Emerge'Naturally'in
  Multi-Agent Dialog}.
\newblock \bibinfo{journal}{{\em arXiv preprint arXiv:1706.08502\/}}
  (\bibinfo{year}{2017}).
\newblock


\bibitem[\protect\citeauthoryear{Lazaridou, Hermann, Tuyls, and
  Clark}{Lazaridou et~al\mbox{.}}{2018}]%
        {lazaridou2018emergence}
\bibfield{author}{\bibinfo{person}{Angeliki Lazaridou},
  \bibinfo{person}{Karl~Moritz Hermann}, \bibinfo{person}{Karl Tuyls}, {and}
  \bibinfo{person}{Stephen Clark}.} \bibinfo{year}{2018}\natexlab{}.
\newblock \showarticletitle{Emergence of linguistic communication from
  referential games with symbolic and pixel input}.
\newblock \bibinfo{journal}{{\em arXiv preprint arXiv:1804.03984\/}}
  (\bibinfo{year}{2018}).
\newblock


\bibitem[\protect\citeauthoryear{Lazaridou, Peysakhovich, and Baroni}{Lazaridou
  et~al\mbox{.}}{2016}]%
        {lazaridou2016multi}
\bibfield{author}{\bibinfo{person}{Angeliki Lazaridou},
  \bibinfo{person}{Alexander Peysakhovich}, {and} \bibinfo{person}{Marco
  Baroni}.} \bibinfo{year}{2016}\natexlab{}.
\newblock \showarticletitle{Multi-agent cooperation and the emergence of
  (natural) language}.
\newblock \bibinfo{journal}{{\em arXiv preprint arXiv:1612.07182\/}}
  (\bibinfo{year}{2016}).
\newblock


\bibitem[\protect\citeauthoryear{LeCun, Bengio, and Hinton}{LeCun
  et~al\mbox{.}}{2015}]%
        {lecun2015deep}
\bibfield{author}{\bibinfo{person}{Yann LeCun}, \bibinfo{person}{Yoshua
  Bengio}, {and} \bibinfo{person}{Geoffrey Hinton}.}
  \bibinfo{year}{2015}\natexlab{}.
\newblock \showarticletitle{Deep learning}.
\newblock \bibinfo{journal}{{\em nature\/}} \bibinfo{volume}{521},
  \bibinfo{number}{7553} (\bibinfo{year}{2015}), \bibinfo{pages}{436}.
\newblock


\bibitem[\protect\citeauthoryear{Lee, Cho, Weston, and Kiela}{Lee
  et~al\mbox{.}}{2017}]%
        {lee2017emergent}
\bibfield{author}{\bibinfo{person}{Jason Lee}, \bibinfo{person}{Kyunghyun Cho},
  \bibinfo{person}{Jason Weston}, {and} \bibinfo{person}{Douwe Kiela}.}
  \bibinfo{year}{2017}\natexlab{}.
\newblock \showarticletitle{Emergent translation in multi-agent communication}.
\newblock \bibinfo{journal}{{\em arXiv preprint arXiv:1710.06922\/}}
  (\bibinfo{year}{2017}).
\newblock


\bibitem[\protect\citeauthoryear{Lerer and Peysakhovich}{Lerer and
  Peysakhovich}{2018}]%
        {lerer2018learning}
\bibfield{author}{\bibinfo{person}{Adam Lerer} {and} \bibinfo{person}{Alexander
  Peysakhovich}.} \bibinfo{year}{2018}\natexlab{}.
\newblock \showarticletitle{Learning Social Conventions in Markov Games}.
\newblock \bibinfo{journal}{{\em arXiv preprint arXiv:1806.10071\/}}
  (\bibinfo{year}{2018}).
\newblock


\bibitem[\protect\citeauthoryear{Littman}{Littman}{1994}]%
        {littman1994markov}
\bibfield{author}{\bibinfo{person}{Michael~L Littman}.}
  \bibinfo{year}{1994}\natexlab{}.
\newblock \showarticletitle{Markov games as a framework for multi-agent
  reinforcement learning}. In \bibinfo{booktitle}{{\em Proceedings of the
  eleventh international conference on machine learning}},
  Vol.~\bibinfo{volume}{157}. \bibinfo{pages}{157--163}.
\newblock


\bibitem[\protect\citeauthoryear{Mnih, Badia, Mirza, Graves, Lillicrap, Harley,
  Silver, and Kavukcuoglu}{Mnih et~al\mbox{.}}{2016}]%
        {mnih2016asynchronous}
\bibfield{author}{\bibinfo{person}{Volodymyr Mnih},
  \bibinfo{person}{Adria~Puigdomenech Badia}, \bibinfo{person}{Mehdi Mirza},
  \bibinfo{person}{Alex Graves}, \bibinfo{person}{Timothy Lillicrap},
  \bibinfo{person}{Tim Harley}, \bibinfo{person}{David Silver}, {and}
  \bibinfo{person}{Koray Kavukcuoglu}.} \bibinfo{year}{2016}\natexlab{}.
\newblock \showarticletitle{Asynchronous methods for deep reinforcement
  learning}. In \bibinfo{booktitle}{{\em International Conference on Machine
  Learning}}. \bibinfo{pages}{1928--1937}.
\newblock


\bibitem[\protect\citeauthoryear{Mordatch and Abbeel}{Mordatch and
  Abbeel}{2017}]%
        {mordatch2017emergence}
\bibfield{author}{\bibinfo{person}{Igor Mordatch} {and} \bibinfo{person}{Pieter
  Abbeel}.} \bibinfo{year}{2017}\natexlab{}.
\newblock \showarticletitle{Emergence of Grounded Compositional Language in
  Multi-Agent Populations}.
\newblock \bibinfo{journal}{{\em arXiv preprint arXiv:1703.04908\/}}
  (\bibinfo{year}{2017}).
\newblock


\bibitem[\protect\citeauthoryear{Nowak and Krakauer}{Nowak and
  Krakauer}{1999}]%
        {nowak1999evolution}
\bibfield{author}{\bibinfo{person}{Martin~A Nowak} {and}
  \bibinfo{person}{David~C Krakauer}.} \bibinfo{year}{1999}\natexlab{}.
\newblock \showarticletitle{The evolution of language}.
\newblock \bibinfo{journal}{{\em Proceedings of the National Academy of
  Sciences\/}} \bibinfo{volume}{96}, \bibinfo{number}{14}
  (\bibinfo{year}{1999}), \bibinfo{pages}{8028--8033}.
\newblock


\bibitem[\protect\citeauthoryear{Paszke, Gross, Chintala, Chanan, Yang, DeVito,
  Lin, Desmaison, Antiga, and Lerer}{Paszke et~al\mbox{.}}{2017}]%
        {paszke2017automatic}
\bibfield{author}{\bibinfo{person}{Adam Paszke}, \bibinfo{person}{Sam Gross},
  \bibinfo{person}{Soumith Chintala}, \bibinfo{person}{Gregory Chanan},
  \bibinfo{person}{Edward Yang}, \bibinfo{person}{Zachary DeVito},
  \bibinfo{person}{Zeming Lin}, \bibinfo{person}{Alban Desmaison},
  \bibinfo{person}{Luca Antiga}, {and} \bibinfo{person}{Adam Lerer}.}
  \bibinfo{year}{2017}\natexlab{}.
\newblock \showarticletitle{Automatic differentiation in pytorch}.
\newblock  (\bibinfo{year}{2017}).
\newblock


\bibitem[\protect\citeauthoryear{Pearl, Glymour, and Jewell}{Pearl
  et~al\mbox{.}}{2016}]%
        {pearl2016causal}
\bibfield{author}{\bibinfo{person}{Judea Pearl}, \bibinfo{person}{Madelyn
  Glymour}, {and} \bibinfo{person}{Nicholas~P Jewell}.}
  \bibinfo{year}{2016}\natexlab{}.
\newblock \bibinfo{booktitle}{{\em Causal inference in statistics: a primer}}.
\newblock \bibinfo{publisher}{John Wiley \& Sons}.
\newblock


\bibitem[\protect\citeauthoryear{Pearson}{Pearson}{1901}]%
        {pearson1901principal}
\bibfield{author}{\bibinfo{person}{Karl Pearson}.}
  \bibinfo{year}{1901}\natexlab{}.
\newblock \showarticletitle{Principal components analysis}.
\newblock \bibinfo{journal}{{\em The London, Edinburgh, and Dublin
  Philosophical Magazine and Journal of Science\/}} \bibinfo{volume}{6},
  \bibinfo{number}{2} (\bibinfo{year}{1901}), \bibinfo{pages}{559}.
\newblock


\bibitem[\protect\citeauthoryear{Sally}{Sally}{1995}]%
        {sally1995conversation}
\bibfield{author}{\bibinfo{person}{David Sally}.}
  \bibinfo{year}{1995}\natexlab{}.
\newblock \showarticletitle{Conversation and cooperation in social dilemmas: A
  meta-analysis of experiments from 1958 to 1992}.
\newblock \bibinfo{journal}{{\em Rationality and society\/}}
  \bibinfo{volume}{7}, \bibinfo{number}{1} (\bibinfo{year}{1995}),
  \bibinfo{pages}{58--92}.
\newblock


\bibitem[\protect\citeauthoryear{Schulman, Levine, Abbeel, Jordan, and
  Moritz}{Schulman et~al\mbox{.}}{2015}]%
        {schulman2015trust}
\bibfield{author}{\bibinfo{person}{John Schulman}, \bibinfo{person}{Sergey
  Levine}, \bibinfo{person}{Pieter Abbeel}, \bibinfo{person}{Michael Jordan},
  {and} \bibinfo{person}{Philipp Moritz}.} \bibinfo{year}{2015}\natexlab{}.
\newblock \showarticletitle{Trust region policy optimization}. In
  \bibinfo{booktitle}{{\em International Conference on Machine Learning}}.
  \bibinfo{pages}{1889--1897}.
\newblock


\bibitem[\protect\citeauthoryear{Smith}{Smith}{1991}]%
        {smith1991honest}
\bibfield{author}{\bibinfo{person}{John~M Smith}.}
  \bibinfo{year}{1991}\natexlab{}.
\newblock \showarticletitle{Honest signalling: The Philip Sidney game.}
\newblock \bibinfo{journal}{{\em Animal Behaviour\/}} (\bibinfo{year}{1991}).
\newblock


\bibitem[\protect\citeauthoryear{Sukhbaatar, Fergus, et~al\mbox{.}}{Sukhbaatar
  et~al\mbox{.}}{2016}]%
        {sukhbaatar2016learning}
\bibfield{author}{\bibinfo{person}{Sainbayar Sukhbaatar}, \bibinfo{person}{Rob
  Fergus}, {et~al\mbox{.}}} \bibinfo{year}{2016}\natexlab{}.
\newblock \showarticletitle{Learning multiagent communication with
  backpropagation}. In \bibinfo{booktitle}{{\em Advances in Neural Information
  Processing Systems}}. \bibinfo{pages}{2244--2252}.
\newblock


\bibitem[\protect\citeauthoryear{Sutton, McAllester, Singh, and Mansour}{Sutton
  et~al\mbox{.}}{2000}]%
        {sutton2000policy}
\bibfield{author}{\bibinfo{person}{Richard~S Sutton}, \bibinfo{person}{David~A
  McAllester}, \bibinfo{person}{Satinder~P Singh}, {and}
  \bibinfo{person}{Yishay Mansour}.} \bibinfo{year}{2000}\natexlab{}.
\newblock \showarticletitle{Policy gradient methods for reinforcement learning
  with function approximation}. In \bibinfo{booktitle}{{\em Advances in neural
  information processing systems}}.
\newblock


\bibitem[\protect\citeauthoryear{Szegedy, Zaremba, Sutskever, Bruna, Erhan,
  Goodfellow, and Fergus}{Szegedy et~al\mbox{.}}{2013}]%
        {szegedy2013intriguing}
\bibfield{author}{\bibinfo{person}{Christian Szegedy},
  \bibinfo{person}{Wojciech Zaremba}, \bibinfo{person}{Ilya Sutskever},
  \bibinfo{person}{Joan Bruna}, \bibinfo{person}{Dumitru Erhan},
  \bibinfo{person}{Ian Goodfellow}, {and} \bibinfo{person}{Rob Fergus}.}
  \bibinfo{year}{2013}\natexlab{}.
\newblock \showarticletitle{Intriguing properties of neural networks}.
\newblock \bibinfo{journal}{{\em arXiv preprint arXiv:1312.6199\/}}
  (\bibinfo{year}{2013}).
\newblock


\bibitem[\protect\citeauthoryear{Wagner, Reggia, Uriagereka, and
  Wilkinson}{Wagner et~al\mbox{.}}{2003}]%
        {wagner2003progress}
\bibfield{author}{\bibinfo{person}{Kyle Wagner}, \bibinfo{person}{James~A
  Reggia}, \bibinfo{person}{Juan Uriagereka}, {and} \bibinfo{person}{Gerald~S
  Wilkinson}.} \bibinfo{year}{2003}\natexlab{}.
\newblock \showarticletitle{Progress in the simulation of emergent
  communication and language}.
\newblock \bibinfo{journal}{{\em Adaptive Behavior\/}} \bibinfo{volume}{11},
  \bibinfo{number}{1} (\bibinfo{year}{2003}), \bibinfo{pages}{37--69}.
\newblock


\bibitem[\protect\citeauthoryear{Williams}{Williams}{1992}]%
        {williams1992simple}
\bibfield{author}{\bibinfo{person}{Ronald~J Williams}.}
  \bibinfo{year}{1992}\natexlab{}.
\newblock \showarticletitle{Simple statistical gradient-following algorithms
  for connectionist reinforcement learning}.
\newblock \bibinfo{journal}{{\em Machine learning\/}} \bibinfo{volume}{8},
  \bibinfo{number}{3-4} (\bibinfo{year}{1992}), \bibinfo{pages}{229--256}.
\newblock


\bibitem[\protect\citeauthoryear{Williams and Peng}{Williams and Peng}{1991}]%
        {williams1991function}
\bibfield{author}{\bibinfo{person}{Ronald~J Williams} {and}
  \bibinfo{person}{Jing Peng}.} \bibinfo{year}{1991}\natexlab{}.
\newblock \showarticletitle{Function optimization using connectionist
  reinforcement learning algorithms}.
\newblock \bibinfo{journal}{{\em Connection Science\/}} \bibinfo{volume}{3},
  \bibinfo{number}{3} (\bibinfo{year}{1991}), \bibinfo{pages}{241--268}.
\newblock


\bibitem[\protect\citeauthoryear{Wittgenstein}{Wittgenstein}{1953}]%
        {wittgenstein1954}
\bibfield{author}{\bibinfo{person}{Ludwig Wittgenstein}.}
  \bibinfo{year}{1953}\natexlab{}.
\newblock \showarticletitle{Philosophical investigations}.
\newblock  (\bibinfo{year}{1953}).
\newblock


\bibitem[\protect\citeauthoryear{Zahavi}{Zahavi}{1975}]%
        {zahavi1975mate}
\bibfield{author}{\bibinfo{person}{Amotz Zahavi}.}
  \bibinfo{year}{1975}\natexlab{}.
\newblock \showarticletitle{Mate selection—a selection for a handicap}.
\newblock \bibinfo{journal}{{\em Journal of theoretical Biology\/}}
  \bibinfo{volume}{53}, \bibinfo{number}{1} (\bibinfo{year}{1975}),
  \bibinfo{pages}{205--214}.
\newblock


\end{thebibliography}

\appendix

\section{Additional algorithms}

For clarity, we provide three additional algorithms related to our paper here.  Algorithm \ref{alg:envstep} describes a single iteration of a matrix communication game (MCG), described in Section \ref{sec:mcgs}. Algorithms \ref{alg:sc} and \ref{alg:ic} describe the calculation of the speaker consistency (SC) and instantaneous coordination (IC), respectively, for both agents. The difference between Algorithms \ref{alg:sc} and \ref{alg:ic} is how the joint probabilities are calculated: for SC, $p(a, m)$ refers to the probability that agent 1 sends message $m$ and then takes action $a$, while for IC it refers to the probability that agent 1 sends message $m$ and then \textit{agent 2} takes action $a$. 

\RestyleAlgo{algoruled}
\CommentSty{}
\begin{algorithm} 
\caption{\texttt{EnvStep} (A single round of an MCG with turn-based communication.) \label{alg:envstep}}
\KwData{Agent policies $\pol_1$ and $\pol_2$, state $S$, payoff matrices $R = (R_1, R_2)$ and observations $O = (o_1, o_2)$.}
\tcp{Randomly select agent $j$ to go first, $j \in \{1, 2\}$.} 
$j \sim \text{Bernoulli}(0.5) + 1$ \\ 
\tcp{Agent $j$ sends a message $m_j$ given observation $o_j$.} 
$m_j \leftarrow \pol_j(m|o_j)$ \\ 
\tcp{Agent $1-j$ sends a message $m_{1-j}$ given observation $o_{1-j}$ and agent $j$'s message.}
$m_{1-j} \leftarrow \pol_{1-j}(m|o_{1-j}, m_j)$ \\  
\tcp{Both agents act simultaneously.}
$a_1 \leftarrow \pol_1(a|o_1, m_1, m_2)$, \hspace{3mm} $a_2 \leftarrow \pol_2(a|o_2, m_1, m_2)$  \\
\tcp{Agents receive reward}
$r_1 \leftarrow R_1[a_1, a_2]$, \hspace{3mm} $r_2 \leftarrow R_2[a_1, a_2]$ \\ 
\Return $(m_1, m_2), (a_1, a_2), (r_1, r_2)$
\end{algorithm}

\RestyleAlgo{algoruled}
\CommentSty{}
\begin{algorithm} 
\caption{Speaker consistency (SC). \label{alg:sc}}
\KwData{Agent policy $\pol_1$, other agent policy $\pol_2$, possible messages $\bar{m}=(m_0, ..., m_{M-1})$, number of test games $T$.}
$\text{SC}_1 = 0$, $\text{SC}_2 = 0$\\
\tcp{Initialize matrices $P$ for tracking (message, action) co-occurrences.} 
$P_1 = 0^{|m| \times |a|}$, $P_2 = 0^{|m| \times |a|}$\\
\For{$i \in \{0, ...,  T - 1\}$} { 
    Generate new state $S$, observations $O$, payoffs $R$. \\
    \tcp{Get agents messages and actions.}
    $(m_1, m_2), (a_1, a_2), (r_1, r_2) \leftarrow \texttt{EnvStep}(S, O, R)$ \\
    \tcp{Update co-occurrence matrices}
    $P_1[m_1, a_1] \mathrel{+}= 1$ \\
    $P_2[m_2, a_2] \mathrel{+}= 1$ \\
}    
\For{$j \in \{1, 2\}$} {
    \tcp{Calculate probabilities from co-occurrence matrix.}
    $p_j(a, m) = P_j / T$ \\
    $p_j(a) = \sum_{m \in \mathcal{A}^m} P_j(a, m) $ \\
    $p_j(m) = \sum_{a \in \mathcal{A}^e} P_j(a, m) $ \\
    \tcp{Calculate mutual information.}
    $\text{SC}_j = \sum_{a \in \mathcal{A}^e} \sum_{m \in \mathcal{A}^m} p_j(a, m) \log \frac{p_j(a, m)}{p_j(a)p_j(m)}$ \\
}
\Return $\text{SC}_1, \text{SC}_2$ 
\end{algorithm}

\RestyleAlgo{algoruled}
\CommentSty{}
\begin{algorithm} 
\caption{Instantaneous coordination (IC). \label{alg:ic}}
\KwData{Agent policy $\pol_1$, other agent policy $\pol_2$, possible messages $\bar{m}=(m_0, ..., m_{M-1})$, number of test games $T$.}
$\text{IC}_1 = 0$, $\text{IC}_2 = 0$\\
\tcp{Initialize matrices $P$ for tracking (message, action) co-occurrences.} 
$P_1 = 0^{|m| \times |a|}$, $P_2 = 0^{|m| \times |a|}$\\
\For{$i \in \{0, ...,  T - 1\}$} { 
    Generate new state $S$, observations $O$, payoffs $R$. \\
    \tcp{Get agents messages and actions.}
    $(m_1, m_2), (a_1, a_2), (r_1, r_2) \leftarrow \texttt{EnvStep}(S, O, R)$ \\
    \tcp{Update co-occurrence matrices. Difference with SC is index of $m, a$.}
    $P_1[m_1, a_2] \mathrel{+}= 1$ \\
    $P_2[m_2, a_1] \mathrel{+}= 1$ \\
}    
\For{$j \in \{1, 2\}$} {
    \tcp{Calculate probabilities from co-occurrence matrix.}
    $p_j(a, m) = P_j / T$ \\
    $p_j(a) = \sum_{m \in \mathcal{A}^m} P_j(a, m) $ \\
    $p_j(m) = \sum_{a \in \mathcal{A}^e} P_j(a, m) $ \\
    \tcp{Calculate mutual information.}
    $\text{IC}_j = \sum_{a \in \mathcal{A}^e} \sum_{m \in \mathcal{A}^m} p_j(a, m) \log \frac{p_j(a, m)}{p_j(a)p_j(m)}$ \\
}
\Return $\text{IC}_1, \text{IC}_2$ 
\end{algorithm}

\section{Additional results}
\label{apdx:results}

\subsection{Fixed matrix SC results} 
In Table \ref{tab:fixedmat}, we show the results of measuring the speaker consistency for various fixed matrices. We have observed that positive signaling does not emerge for fixed payoffs, even in games where it would make sense as a coordination device (e.g. the Battle of the Sexes). Since the agents are always playing the same payoff, they are able to adapt to the action of the opponent, and find a deterministic equilibrium (or cycle through actions for competitive games). 

\begin{table}[h!]
    \centering
    \begin{tabular}{c c c} \toprule
        \textbf{Experiment} & \textbf{Reward} & \textbf{SC} \\ \hline
         Prisoner's Dilemma & 1.001 $\pm$ 0.000   &  0.000 $\pm$ 0.000 \\ 
         Battle of the Sexes & 1.500 $\pm$ 0.316  & 0.000 $\pm$ 0.000  \\ 
         Matching Pennies & 0.000 $\pm$ 0.512  &  0.011 $\pm$ 0.007 \\ 
         Random 2x2 payoff & 1.286 $\pm$ 1.804  & 0.000 $\pm$ 0.000  \\ 
         Random 4x4 payoff & 2.896 $\pm$ 1.721  & 0.001 $\pm$ 0.003  \\ 
         Random 8x8 payoff & 3.997 $\pm$ 1.353  & 0.015 $\pm$ 0.025  \\
         \bottomrule
    \end{tabular}
    \caption{Average reward and speaker consistency for some popular MCGs (see Table \ref{tab:mat_ex} for payoffs), as well as 20 different random fixed payoffs. }
    \label{tab:fixedmat}
\end{table}

\subsection{Reward and SC results on iterated MCGs}
\begin{figure*}[h]
    \centering
    \includegraphics[width=0.45\textwidth]{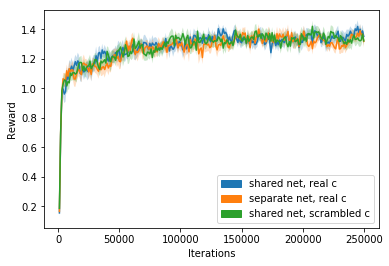}
    \includegraphics[width=0.45\textwidth]{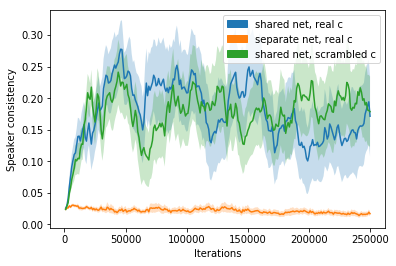}
    \includegraphics[width=0.45\textwidth]{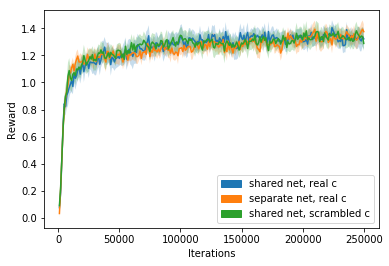}
    \includegraphics[width=0.45\textwidth]{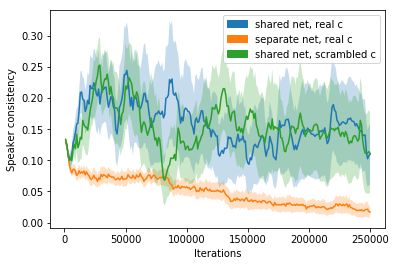}
    \includegraphics[width=0.45\textwidth]{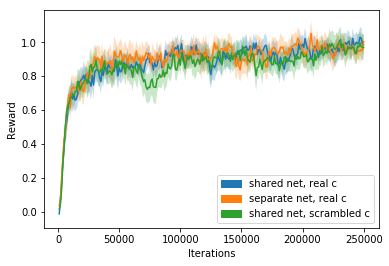}
    \includegraphics[width=0.45\textwidth]{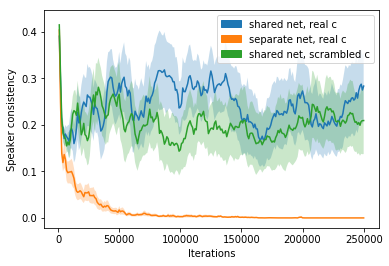}
    \caption{Results for the iterated setting using A2C. Plots of the reward (left) and speaker consistency (right), for the 2x2 (top), 4x4 (middle), and 8x8 (bottom) sized payoff matrix. }
    \label{fig:iterated}
\end{figure*}

The results in our paper are not limited to the non-iterated case. To show this, we run experiments on an iterated version of our environment, using the A2C algorithm \cite{mnih2016asynchronous}. We keep the policy architectures the same, except we give each agent a memory of the previous 5 rounds (the actions and the messages of both agents), which is concatenated to the input at each round. We increase the value of the discount factor $\gamma$ to 0.9, from 0. Changing our REINFORCE algorithm with a learned baseline to A2C requires changing the way $Q^{\pol}$ is estimated; instead of using the next reward, we use the $n$-step return, with $n=5$ (see \cite{mnih2016asynchronous} for details). The results are shown in \ref{fig:iterated}. We can see that the same general trend is present: there is positive SC in the randomized $R$ setting, even when scrambling the messages $c$; however, the SC disappears when a different network is used to produce the messages. 

\subsection{Full metric results for randomized $R$} In Table \ref{tab:random_r_stats}, we show results for the speaker consistency, policy entropy, and context independence in the randomized $R$ setting. In Table \ref{tab:resultslong} we show the reward and speaker consistency evaluated over the last 10,000 games of training for the ablation studies performed in the paper. 

\begin{table*}[]
    \centering
    \begin{tabular}{c c c c c c c}
    \toprule
         & \multicolumn{2}{c}{\textbf{2 $\times$ 2 payoffs}} & \multicolumn{2}{c}{\textbf{4 $\times$ 4 payoffs}} & \multicolumn{2}{c}{\textbf{8 $\times$ 8 payoffs}}  \\
        \textbf{Metrics} & Real $c$ & Random $c$ & Real $c$ & Random $c$ &Real $c$ & Random $c$  \\ \hline
        SC          & 0.187 $\pm$ 0.032  & 0.001 $\pm$ 0.001 &  0.476 $\pm$ 0.063 & 0.008 $\pm$ 0.001 & 0.541 $\pm$ 0.055 & 0.016 $\pm$ 0.001 \\
        $H(\pol^c)$ &  1.116 $\pm$ 0.044 & 1.386 $\pm$ 0.001 & 0.934 $\pm$ 0.087 & 1.789 $\pm$ 0.001 & 0.855 $\pm$ 0.080 & 2.298 $\pm$ 0.001 \\
        CI          & 0.408 $\pm$ 0.044  & 0.137 $\pm$ 0.001 & 0.369 $\pm$ 0.040 & 0.057 $\pm$ 0.001 & 0.182 $\pm$ 0.016 & 0.021 $\pm$ 0.001  \\
        \bottomrule
    \end{tabular}
    \caption{Various metrics calculated in the random $R$ setting described in Section \ref{sec:5.1}. Results are calculated on 1000 test games over 10 random seeds. Averages are shown along with the standard error. }
    \label{tab:random_r_stats}
\end{table*}

\begin{table*}[]
    \centering
    \begin{tabular}{c c c c c c c} \toprule
    \textbf{Training} & \multicolumn{2}{c}{\textbf{2x2 payoff}} & \multicolumn{2}{c}{\textbf{4x4 payoff}} & \multicolumn{2}{c}{\textbf{8x8 payoff}} \\ 
    \textbf{setting} & Reward & SC & Reward & SC & Reward & SC  \\ \hline
        
       Scrambled $c$ &  1.42 $\pm$ 0.08 & 0.20 $\pm$ 0.08  & 1.60 $\pm$ 0.07 & 0.49 $\pm$ 0.11  & 1.13 $\pm$ 0.11 & 0.60 $\pm$ 0.19  \\
       Separate $c$ net &  1.41 $\pm$ 0.08 & 0.03 $\pm$ 0.004  & 1.58 $\pm$ 0.12 & 0.12 $\pm$ 0.02  & 1.12 $\pm$ 0.13 & 0.02 $\pm$ 0.03  \\
        No $c$ training &  1.39 $\pm$ 0.08 & 0.17 $\pm$ 0.07  & 1.65 $\pm$ 0.13 & 0.43 $\pm$ 0.05  & 1.07 $\pm$ 0.15 & 0.69 $\pm$ 0.10  \\ \hline
        Default &  1.43 $\pm$ 0.07 & 0.20 $\pm$ 0.08  & 1.63 $\pm$ 0.09 & 0.51 $\pm$ 0.19  & 1.12 $\pm$ 0.09 & 0.54 $\pm$ 0.18  \\
        \bottomrule
    \end{tabular}
    \caption{The reward and speaker consistency values for different training settings across various matrix sizes. `Scrambled $c$' is where the messages are replaced by a random message before being observed, `Separate $c$ net' is where the action and message networks have no shared parameters, and `No $c$ training' is where $\lambda_{comm}=0$. Error shown is twice the standard error.}
    \label{tab:resultslong}
\end{table*}

\subsection{Result plots}
\begin{figure*}
    \centering
    \includegraphics[width=0.32\linewidth]{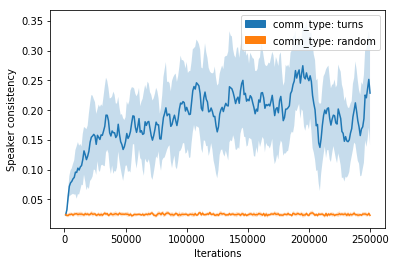}
    \includegraphics[width=0.32\linewidth]{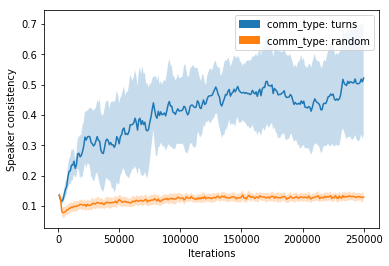}
    \includegraphics[width=0.32\linewidth]{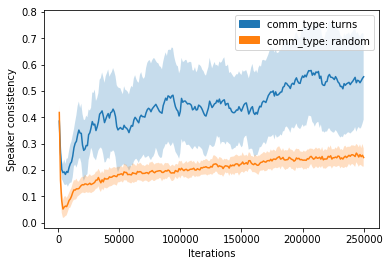}
    \caption{When given the ability to communicate, agents' messages provide information about their subsequent action (from left to right: with 2x2, 4x4, and 8x8 payoff matrices). However, agents don't do this if the message is randomly generated, as would be the case in a correlated equilibrium.}
    \label{fig:comtype}
\end{figure*}

In Figure \ref{fig:comtype}, we plot the SC for various matrix sizes in the randomized $R$ setting. 

\section{Hyperparameter details}

We provide a table of the hyperparameter values below. For all the plots and tables in our paper, unless otherwise specified, we average across 10 random seeds, and plot twice the standard error. For tables where the numbers are calculated on test games after the policy has been trained, we test for 1000 games per seed, for a total of 10,000 games.  The exception for this is when measuring the causal influence in Table \ref{tab:cic}, where we use 5 random seeds.

\begin{table*}
    \centering
    \begin{tabular}{c|c}
        Hyperparameter & Value  \\ \hline
        $\lambda_{comm}$ & 0.1 \\
        $\lambda_{ent}$ & 0.01 \\
        $\lambda_{v}$ & 0.1 \\
        Learning rate & 0.005 \\
        Batch size & 64 \\
        Reward discount ($\gamma$) & 0 (not iterated) \\
        Number of hidden units in policy network & (40, 60, 100) for matrix sizes of (2x2, 4x4, 8x8) \\
        Number of episodes & 250,000 \\
        Optimizer & ADAM \cite{kingma2014adam} with default params \\
    \end{tabular}
    \caption{Hyperparameters}
    \label{tab:hyperparams}
\end{table*}

\section{Examples of matrix games}

In Table \ref{tab:mat_ex}, we provide three examples of popular matrix games. Generally, agents playing this game will act simultaneously, and receive a reward according to the payoff matrix. In general, the agents receive different rewards from each other. We augment these games with a  communication channel, where agents take turns sending discrete one-hot vectors to each other.  For most of the experiments in our paper, the payoff matrices are randomly generated according to a Gaussian distribution with mean 0 and variance 3.

\begin{table*}
\begin{subtable}{.3\linewidth}
\centering
\begin{tabular}{ c| c c}
  & Opera & Fight \\ \hline
 Opera & 2, 1 & 0, 0\\
  Fight & 0, 0 & 1, 2 \\
\end{tabular}%
\caption{Battle of the sexes}
\end{subtable}
\begin{subtable}{.3\linewidth}
\centering
\begin{tabular}{ c |c c}
   & Stag & Rabbits \\ \hline
 Stag & 1, -1 & -1, 1 \\
   Rabbits & -1, 1 & 1, -1 \\
\end{tabular}%
\caption{Matching Pennies}
\end{subtable}
\quad
\begin{subtable}{.3\linewidth}
\centering
\begin{tabular}{ c| c c}
   & Cooperate & Defect \\ \hline
Cooperate & 3, 3 & 0, 4 \\
   Defect & 4, 0 & 1, 1\\
\end{tabular}%
\caption{Prisoner's Dilemma}
\end{subtable}
 \caption{Three popular matrix games, to which we add a communication channel. The tuple of rewards ($r_1$, $r_2$) indicates the rewards for agent 1 and agent 2, respectively. }
    \label{tab:mat_ex}
\end{table*}

\end{document}